\pdfoutput=1
\documentclass[10pt,twocolumn,letterpaper]{article}

\usepackage[pagenumbers]{cvpr}

\usepackage{microtype}

\renewcommand{\paragraph}[1]{\par\vspace{.05em}\noindent\textbf{#1}\hspace{.55em}\ignorespaces}

\usepackage[T1]{fontenc}
\usepackage{xspace}
\usepackage{multirow}
\usepackage{tcolorbox}
\tcbuselibrary{breakable}
\usepackage{pmboxdraw}
\usepackage{placeins}
\usepackage{float}
\newtcolorbox{promptbox}[1]{breakable,lines before break=5,before upper={\widowpenalty=10000 \clubpenalty=10000 },colback=gray!4,colframe=gray!55,boxrule=0.5pt,arc=1.5pt,left=4pt,right=4pt,
  top=3pt,bottom=3pt,toptitle=1pt,bottomtitle=1pt,fonttitle=\small\bfseries,coltitle=black,colbacktitle=gray!14,
  fontupper=\small,before skip=5pt,after skip=5pt,title={#1}}
\newcommand{\bench}{S2Bench\xspace}

\newcommand{\harness}[1]{{\fontsize{6}{7}\selectfont #1}}
\newcommand{\nregisteredrefs}{65}
\newcommand{\nscenes}{128}

\newcommand{\nexpertrefs}{63}

\definecolor{cvprblue}{rgb}{0.21,0.49,0.74}
\usepackage[pagebackref,breaklinks,colorlinks,allcolors=cvprblue]{hyperref}

\def\confName{CVPR}
\def\confYear{2027}

\title{From Point Clouds to Engineering Models: Benchmarking Agentic 3D Reconstruction in the Built Environment}
\author{Changyu Du\textsuperscript{1} \quad Stavros Nousias\textsuperscript{1} \quad Syed Salman Ahmed\textsuperscript{1}\\
Kai Dietmair\textsuperscript{1} \quad Mohammad Alrabab'h\textsuperscript{1} \quad Andr\'e Borrmann\textsuperscript{1}\\[0.5em]
\textsuperscript{1}Technical University of Munich\\
{\tt\small changyu.du@tum.de}}
\hypersetup{pdftitle={From Point Clouds to Engineering Models: Benchmarking Agentic 3D Reconstruction in the Built Environment},pdfauthor={Changyu Du, Stavros Nousias, Syed Salman Ahmed, Kai Dietmair, Mohammad Alrabab'h, Andr\'e Borrmann}}

\begin{document}
\maketitle

\begin{abstract}
Coding agents can translate 3D measurements into CAD geometry, but reconstructing engineering models for the built environment demands semantic interpretation and consistent construction constraints beyond surface fitting. We formulate scene-level Scan-to-BIM (Building Information Modeling) as an agentic 3D coding task and introduce \bench, comprising public and proprietary scans of 128 real-world scenes across five domains, paired with expert-curated reference models spanning diverse spatial scales. A shared framework lets agents measure scans in Python, build class-labeled components in a live CAD/BIM kernel, and iteratively refine their code. We also propose parallel asynchronous collaboration, where a lead delegates versioned tasks, continues modeling while workers execute, and integrates revision-bound proposals. Evaluating five agents and a collaborative team across 1,028 reconstructions shows that scan-to-model residuals directly guide geometric repair, whereas inferring semantic classes and construction relations requires joint geometric, visual, and engineering reasoning that agents struggle to ground and verify. Consequently, the leading agent fits scan geometry closely ($96\%$ coverage within $5\,\text{cm}$) yet reaches only $63\%$ mIoU, and no agent passes all engineering checks on more than $38\%$ of scenes. Coverage and engineering conformance drop with scene scale, whereas running a smaller model as a team improves all three axes with greater coverage gains on larger scenes. The project page is available at \href{https://tumcms.github.io/S2Bench.github.io/}{\nolinkurl{tumcms.github.io/S2Bench.github.io}}.
\end{abstract}

\begin{figure}[t]
  \centering
  \includegraphics[width=\linewidth]{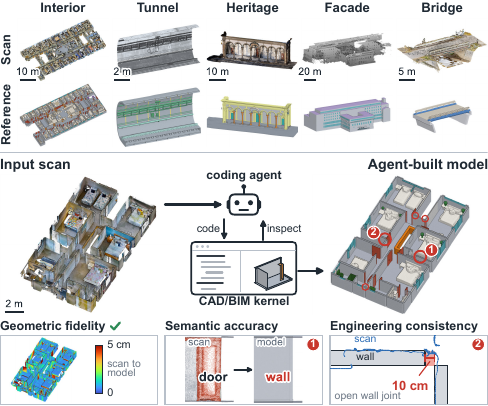}
  \caption{\textbf{Overview of \bench.} \bench pairs scans of 128 scenes across five built-environment domains and spatial scales with expert-curated reference models (top). A coding agent builds a component model by writing and inspecting code in a live CAD/BIM kernel (middle). Despite high geometric fidelity, agent-built models can fail semantic accuracy (1: door as wall) or engineering consistency (2: open wall joint).}
  \label{fig:teaser}
\end{figure}

\section{Introduction}
\label{sec:intro}

While 3D point clouds capture the surfaces of built structures, engineering models organize these observations into discrete components with explicit geometric, semantic, and relational meaning. Simply fitting primitives to a scan does not distinguish a structural beam, anchor it to its supports, or preserve a navigable doorway. Recovering these decisions is central to Scan-to-BIM (Building Information Modeling), which converts scans into structured component models for dimensional verification, retrofit planning, and other engineering applications. In practice, modeling still relies heavily on experts, while automated pipelines largely target specific component families and scene types~\cite{zbirovsky2025cloud2bim,chamseddine2026bimstruct3d}.

Coding agents offer a way to make these reconstruction decisions at execution time~\cite{wang2024codeact}. Rather than relying on fixed heuristics, an agent can measure a point cloud, instantiate components via code, and refine the model through visual and quantitative inspection. Recent studies apply this closed-loop approach to indoor rooms~\cite{huang2026literealityagent}, everyday objects~\cite{liu20263dharnessbench}, and CAD tasks~\cite{mallis2025cadassistant}. Extending it across the built environment requires adapting strategies to diverse structures and scales, from indoor partitions to long-span bridges and historic vaults, within a bounded budget while preserving component meaning and valid construction relations.

Because surface proximity alone misses unseen contact faces, spurious fragments, and semantic swaps, usable engineering models test capabilities existing benchmarks do not measure: inferring implicit semantic classes and construction constraints from noisy scans rather than generating from specifications~\cite{yang2026dreamhouse,dong2026muse} or fitting isolated objects~\cite{liu20263dharnessbench}. We therefore evaluate three axes jointly (\cref{fig:teaser}): geometric fidelity to the scan and reference surfaces, semantic accuracy of component classes, and engineering consistency of solids, joints, supports, clashes, and openings.

To this end, we introduce \textbf{\bench}, a benchmark framing scene-level Scan-to-BIM as an agentic 3D coding task across 128 real-world scans (122 public from 13 datasets and six proprietary bridges) in five building domains, paired with expert-curated references. Given a scan and class vocabulary, agents choose measurement, decomposition, and modeling strategies in a shared CAD/BIM environment. To overcome single-agent budget and context limits on large scenes, we also propose parallel asynchronous reconstruction: a lead delegates sub-tasks, continues modeling, and merges versioned proposals into one model, complementing staged workflows~\cite{pan2026multiagent}.

Across 1,028 reconstructions, performance drops sharply from geometry to semantics and relations: the top agent reaches 95.6\% scan coverage yet only 63.2\% mIoU, and no agent passes all engineering checks on more than 37.5\% of scenes. Scan-to-model residuals guide surface repair, whereas flush components and construction errors leave little distance residual and evade self-referential checks. Chasing tighter scan fit with dense meshes can even hurt structural consistency: GPT-6.1 Sol leads in coverage and mIoU, while Claude Opus 5.5 favors clean solids and leads in engineering conformance. Larger scenes widen gaps, whereas running the smaller GPT-6.1 Terra as a team lifts coverage from 72.7\% to 89.8\% and nearly doubles engineering conformance, with the steepest gains on large scenes.
In summary, our contributions are:
\begin{itemize}[nosep,leftmargin=*]
  \item \textbf{\bench}, a 128-scene benchmark across five built-environment domains and diverse spatial scales with expert-curated reference models for agentic Scan-to-BIM.
  \item A shared CAD/BIM environment for closed-loop scan measurement and code execution, plus a parallel asynchronous multi-agent strategy with versioned delegation and interface-checked proposal merging.
  \item A three-axis evaluation protocol measuring geometric fidelity to scans and references surfaces, semantic accuracy, and conformance to nine engineering rules.
  \item Analysis of 1,028 reconstructions, 640 single-agent trajectories, and 22-scene ablations isolating the semantic and relational bottlenecks behind high scan coverage and show when parallel agents recover scale losses.
\end{itemize}

\section{Related Work}
\label{sec:related}

\begingroup
\setlength{\parskip}{0pt}
\noindent\textbf{3D modeling with code.}
Programs make 3D geometry editable and record how it was built. Learned methods predict programs for indoor layouts~\cite{avetisyan2024scenescript,mao2025spatiallm} and CAD parts~\cite{rukhovich2024cadrecode} from task-specific data. Coding agents instead execute, inspect, and revise code for scene composition, image-based reconstruction, and CAD~\cite{hu2024scenecraft,yin2026viga,mallis2025cadassistant}. Text2BIM~\cite{du2026text2bim} and LiteReality-Agent~\cite{huang2026literealityagent} add rule-based revision and collision or support feedback. We study how agents ground such programs in measured surfaces to recover components, classes, and construction relations.

\noindent\textbf{Scan-to-BIM and agentic reconstruction.}
Scan-to-BIM methods recover surfaces, volumetric walls, and floor plans from geometric assumptions~\cite{nan2017polyfit,ochmann2019volumetric,yue2023roomformer}, and later systems handle topology correction, non-orthogonal walls, bridge parameters, and MEP components~\cite{chamseddine2026bimstruct3d,zbirovsky2025cloud2bim,fang2025scantobimtosim,wang2024omniscan2bim}, though their routines remain tailored to specific asset families. Agentic approaches instead make modeling decisions during execution, reconstructing rooms from video~\cite{lin2026grase}, exteriors from images~\cite{pantojarosero2026astralod3}, or indoor BIMs from floor plans and panoramas~\cite{castano2026pano2bim}. Agent2BIM~\cite{wang2026agent2bim} studies capture evidence with one agent on two scenes, and Pan et al.~\cite{pan2026multiagent} use a staged multi-agent workflow. While these studies show feasibility in single domains, we compare agents across five domains and spatial scales under one task, and evaluate asynchronous collaboration on a shared model.

\noindent\textbf{Benchmarks and datasets.}
3D agent benchmarks cover scene editing and object modeling~\cite{gu2025blendergym,gao20263dcodebench}, reconstruction under controlled evidence~\cite{liu20263dharnessbench}, and generation from specifications~\cite{fang2026blenderworld}. CAD benchmarks test GUI workflows and assembly interfaces~\cite{dong2026cadworld,dong2026muse}, and DreamHouse~\cite{yang2026dreamhouse} checks timber frames. Unlike specification-driven generation or single-object tasks, \bench tests inferring implicit structural constraints from raw scans across scales without given connectivity graphs. Perception datasets provide scans and labels~\cite{armeni2016s3dis,wysocki2022tumfacade,kellner2026semanticbridge}, and BIMNet and DeKH add authored BIMs~\cite{bimnet,chamseddine2026bimstruct3d}, yet none pairs heterogeneous built-environment scans with unified scan fidelity, bidirectional reference $F_1$, and cross-domain engineering conformance checks. Software benchmarks report limited team gains over strong single agents~\cite{khatua2026cooperbench,kim2025scaling}, whereas we measure collaboration where local edits must satisfy shared physical interfaces.

\par\endgroup

\section{\bench}
\label{sec:bench}

\subsection{Task and scenes}
\bench covers five domains differing in component vocabulary and geometric demands (\cref{tab:bench}), with 122 public scans from 13 datasets and six held-out proprietary bridge scans. For domain balance, we randomly subsample three large sources: Rohbau3D (6 of 504, one per site), STSD (10 of 264 across three tunnel cross-sections), and 3DSES (20 of 39 across annotation tiers). All other sources contribute their labeled scenes, retaining only scenes with a reference model. \nregisteredrefs\ references come from source datasets, and five domain experts built \nexpertrefs\ new references (\cref{sec:reference_models}). Scans are voxel-downsampled to 1\,cm in a metric Z-up frame with available color and intensity. For controlled ablations and repeat runs, we designate a 22-scene stratified subset (\texttt{core22}) spanning every public source and domain.

Given a scan $X$ and a component vocabulary $\mathcal V$, an agent $\pi$ writes a construction program $\Pi$, and executing $\Pi$ in the modeling environment $\mathcal E$ yields the model $M$:
\begin{equation}
 \Pi=\pi(X,\mathcal V),\qquad M=\mathcal E(\Pi)=\{(g_i,c_i)\}_{i=1}^{n}.
\label{eq:task}
\end{equation}
Each component has geometry $g_i$ in the metric frame of $X$, class $c_i\in\mathcal V$, and a stable identity. Within a 45-minute budget, the agent chooses its measurements, decomposition, and operations in \cref{sec:method} to output $\Pi$, a native model, and an IFC export. \Cref{sec:eval} scores $M$ against $X$, point labels, and the reference model, which stay withheld except in the label-assistance study (\cref{tab:interv} and Supp.\ \cref{sec:supp_interventions}).

\begin{table}[t]
  \centering
  \footnotesize
  \setlength{\tabcolsep}{0.6pt}
  \renewcommand{\arraystretch}{0.92}
  \begin{tabular*}{\linewidth}{@{\extracolsep{\fill}}llrr@{}}
      \toprule
      Dataset & Scene type & Points (M) & Extent (m) \\
      \midrule
      \multicolumn{4}{@{}l}{\textcolor[HTML]{4C78A8}{\textbf{Interiors (73)}}} \\
      3DSES~\cite{merizette2025_3dses} & Academic buildings & 0.6--1.4 & 8--29 \\
      BIMNet~\cite{bimnet} & Mixed interiors & 0.7--23.3 & 9--100 \\
      CRAS Labs~\cite{abreu2023cras} & Laboratories & 20.4 & 31 \\
      DeKH~\cite{chamseddine2026bimstruct3d} & Hospital & 22.0--35.7 & 27--40 \\
      LivingBIM~\cite{alsuhaibani2025livingbim} & Academic buildings & 2.5--17.0 & 9--53 \\
      Rohbau3D~\cite{rauch2025rohbau3d} & Construction, renovation & 1.3--9.4 & 23--71 \\
      S3DIS~\cite{armeni2016s3dis} & Offices, education & 11.7--27.5 & 39--54 \\
      UGD~\cite{yue2025ugd} & Underground garages & 5.2--11.0 & 25--49 \\
      \midrule
      \multicolumn{4}{@{}l}{\textcolor[HTML]{28796F}{\textbf{Bridges (34)}}} \\
      Japan RC Bridges~\cite{chun2025japanbridges} & Single span & 2.9--16.7 & 11--47 \\
      SemanticBridge~\cite{kellner2026semanticbridge} & Single, multiple spans & 4.6--19.0 & 62--127 \\
      Weilheim$^\dagger$ & Single, multiple spans & 3.4--17.9 & 58--166 \\
      \midrule
      \multicolumn{4}{@{}l}{\textcolor[HTML]{9A6B16}{\textbf{Tunnels (10)}}} \\
      STSD~\cite{cui2024stsd} & \begin{tabular}[t]{@{}l@{}}Circular, horseshoe,\\quasi-rectangular\end{tabular} & 2.1--2.8 & 14--18 \\
      \midrule
      \multicolumn{4}{@{}l}{\textcolor[HTML]{A85548}{\textbf{Facades (7)}}} \\
      TUM-FA\c{C}ADE~\cite{wysocki2022tumfacade} & University campus buildings & 5.3--70.3 & 52--189 \\
      \midrule
      \multicolumn{4}{@{}l}{\textcolor[HTML]{736494}{\textbf{Heritage (4)}}} \\
      Cultural Heritage~\cite{pellis2025heritage} & Arcaded historic buildings & 7.1--18.6 & 35--82 \\
      \bottomrule
    \end{tabular*}
  \caption{\textbf{Sources and scales.} Domain scene counts are in parentheses. Ranges give point counts (1\,cm voxelized) and bounding-box diagonal extent. $^\dagger$Self-collected, unpublished.}
  \label{tab:bench}
\end{table}

\subsection{Reference models}
\label{sec:reference_models}
Five domain experts register existing references to the scan frame and build the \nexpertrefs\ new references by revising initial Sol reconstructions with annotated point clouds. Revisions are substantial because starting meshes overfit scan noise and are rebuilt as discrete components. Joint review of initial models surfaces issues missed by scan distance: mislabeled, missing, or spurious components, plus connection, gap, and overlap errors (Supp.\ \cref{tab:expert_comments}), motivating our semantic and engineering axes (\cref{eq:semantic_main,eq:ecr}). All five experts cross-review the final references before calibrating the checker (\cref{sec:eval}). Scan coverage and mIoU do not use references, and Sol's $F_1^{\mathrm{surf}}$ is scored only on separate runs not used as starting models (Supp.\ \cref{sec:supp_reference}).

\subsection{Evaluation protocol}
\label{sec:eval}
We score reconstructed models on geometric fidelity, semantic accuracy, and engineering consistency.

\paragraph{Geometric fidelity.}
Let $X_{\mathrm{eval}}\subseteq X$ denote scan points with non-ignored task classes, and $d(x,M)$ the distance from $x\in X_{\mathrm{eval}}$ to the nearest component surface of $M$. Coverage $R_\tau$ is the fraction of $X_{\mathrm{eval}}$ within tolerance $\tau$ of $M$. Against reference $M^{\mathrm{ref}}$, surface precision $P^{\mathrm{surf}}$ is the surface-area fraction of $M$ within $\tau$ of $M^{\mathrm{ref}}$, and recall $R^{\mathrm{surf}}$ is that of $M^{\mathrm{ref}}$ within $\tau$ of $M$:
\begin{gather}
 \mbox{\footnotesize$\displaystyle R_\tau{=}\frac{1}{|X_{\mathrm{eval}}|}\!\sum_{x\in X_{\mathrm{eval}}}\!\mathbf{1}[d(x,M)\leq\tau],\, F_1^{\mathrm{surf}}{=}\frac{2P^{\mathrm{surf}}R^{\mathrm{surf}}}{P^{\mathrm{surf}}+R^{\mathrm{surf}}}.$}
\label{eq:geometry_main}
\end{gather}
R5 reports 5\,cm coverage as a percentage. Scan-to-model measures~\cite{ersoz2026confidence} ignore unseen back and contact faces and do not penalize extra geometry. Reference surface $F_1$ ($F_1^{\mathrm{surf}}$) closes this gap by penalizing extra and missing area at $\tau=10$\,cm, the ISPRS modelling benchmark tolerance~\cite{khoshelham2021isprs}, over the shared task region and reference-supported classes (zero when no area matches).

\paragraph{Semantic accuracy.}
We transfer component classes to scan points using nearby surfaces and bounded solid containment. For each task class $c$, $\mathrm{TP}_c$, $\mathrm{FP}_c$, and $\mathrm{FN}_c$ count correctly assigned, incorrectly assigned, and missed points. Mean intersection-over-union is
\begin{equation}
 \mathrm{mIoU}=\frac{1}{|\mathcal C|}\sum_{c\in\mathcal C}
 \frac{\mathrm{TP}_c}{\mathrm{TP}_c+\mathrm{FP}_c+\mathrm{FN}_c}.
 \label{eq:semantic_main}
\end{equation}
The mean includes task classes present in the annotation or prediction. Missing, unclassified, and wrong-class predictions remain errors.

\paragraph{Engineering consistency.}
Distance and overlap metrics cannot tell whether components connect, rest on supports, or leave openings clear. We therefore develop a model checker~\cite{nikoohemat2021consistency} with nine rules (\cref{tab:rules}) that target the connection, gap, and overlap errors flagged by the experts (Supp.\ \cref{tab:expert_comments}). Three rules check component integrity (T01--T03), four check connection and support (T04--T07), and two check interference and opening clearance (T08--T09). Each rule is a geometric predicate with declared tolerances (Supp.\ \cref{sec:supp_eng}).

Rules and tolerances are shared, whereas applicability varies by scene: tunnels have no columns, for instance, and bridge girders rest on piers. Each scene therefore uses a frozen protocol specifying its task region, required component types, and required junctions or supports, mostly drawn from domain and dataset defaults. Protocols are calibrated solely on the reference models, each of which passes every check, and verified with injected defects and an expert review of sampled verdicts (Supp.\ \cref{sec:supp_eng}).

At scoring time, the checker reads only the agent's model $M$, meaning its component geometries $g_i$ and classes $c_i$, and the frozen protocol, without scan labels, reference models, or post-hoc geometry repair. For each applicable unit, such as a wall interface or beam support, it returns PASS or FAIL with the components involved, testing geometric construction rather than structural capacity or code compliance.

\begin{table}[t]
  \centering\footnotesize
  \setlength{\tabcolsep}{3pt}
  \renewcommand{\arraystretch}{0.95}
  \begin{tabular}{@{}lp{0.87\linewidth}@{}}
    \toprule
    Rule & Requirement within the declared scene scope \\
    \midrule
    T01 & Components satisfy their declared geometric representation. \\
    T02 & No duplicated components or material regions. \\
    T03 & Components preserve declared intra-object connectivity without stray fragments. \\
    \addlinespace[1pt]
    T04 & Adjacent walls connect at their required interfaces. \\
    T05 & Beams and girders meet their required support conditions. \\
    T06 & Column and pier bases connect to permitted supports. \\
    T07 & Structural members have a support path to a declared root. \\
    \addlinespace[1pt]
    T08 & No material interference outside permitted joint regions. \\
    T09 & Opening clearances remain free of host obstruction. \\
    \bottomrule
  \end{tabular}
  \caption{\textbf{Nine engineering rules.} Predicates and tolerances are detailed in Supp.\ \cref{sec:supp_eng}.}
  \label{tab:rules}
\end{table}

For scene $s$, let $p_{sr}$ and $f_{sr}$ count passing and failing units on rule $r$, and let $f_s=\sum_r f_{sr}$. Over all scenes $\mathcal S$ and shared applicable scenes $\mathcal S_r$, the Engineering Conformance Rate (ECR) and Rule Conformance Rate (RCR-T01 to RCR-T09) give the share (reported as percentages) of scenes with zero failing units:
\begin{equation}
\mathrm{ECR}=\frac{1}{|\mathcal S|}\!\sum_{s}\!\mathbf{1}[f_{s}{=}0],\ \mathrm{RCR}_r=\frac{1}{|\mathcal S_r|}\!\sum_{s}\!\mathbf{1}[f_{sr}{=}0].
\label{eq:ecr}
\end{equation}
Both are binary per scene on valid, non-empty checks, so a model counts as conforming for a relation only if no unit of it fails. Failed or invalid submissions, models exceeding the checker's resource limits (2.5M triangles or 3,600\,s), missing required components, and empty required scopes count as non-conforming. For small-cohort ablations (\cref{tab:interv}), we also compute the nine-rule mean of the unit-level pass rate $\overline q_r$, the mean over $\mathcal S_r$ of $q_{sr}=p_{sr}/(p_{sr}+f_{sr})$ (Supp.\ \cref{sec:supp_eng_results}).
Scan and semantic rates give scenes equal weight, with zero for failed or invalid submissions. Exact definitions, point-transfer settings, and checker implementation appear in Supp.\ \cref{sec:supp_scoring,sec:supp_reference,sec:supp_eng}.

\section{Agentic Reconstruction}
\label{sec:method}

\begin{figure*}[t]
  \centering
  \includegraphics[width=\linewidth]{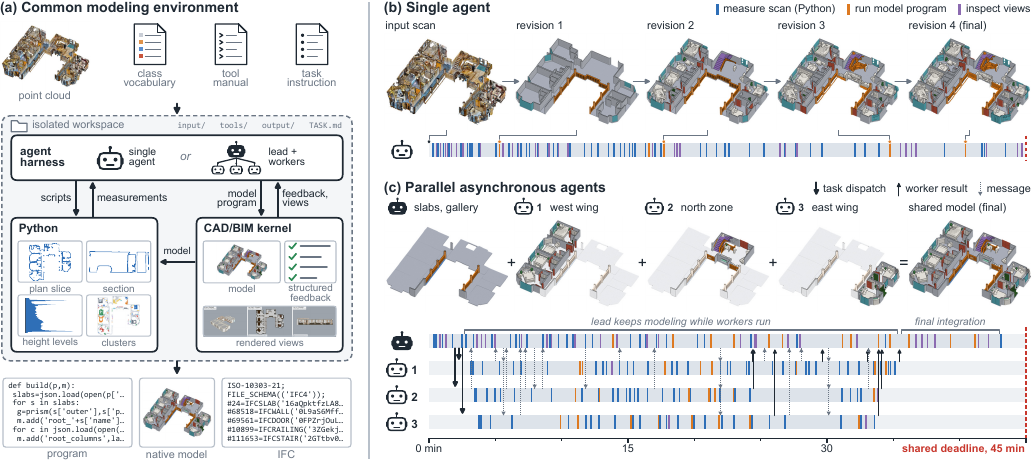}
  \caption{\textbf{Reconstruction framework.} (a) Within an isolated workspace, a harness runs one agent or a team to measure scans in Python and iterate code in Rhino~8, receiving structured feedback (execution errors, validity checks, and agent-written validation) and rendered views before exporting the program, model, and IFC. (b) Single-agent timeline and revisions. (c) Team run on the same scan: the lead (solid icon) delegates regions to three workers, keeps modeling, and merges their parts under a shared 45-minute deadline.}
  \label{fig:method}
\end{figure*}

\subsection{Modeling environment}
\label{sec:environment}
Reconstruction combines numerical scan access with a persistent model via three components in $\mathcal E$ (\cref{fig:method}a). A harness drives the tool loop, Python exposes the scan and exported meshes for geometric queries, and a CAD/BIM backend executes $\Pi$ to maintain $M$. Each execution returns rendered views (optionally overlaid on the scan) and structured feedback combining execution errors, automatic checks (solid closure, duplicates, etc.), and agent-written \texttt{validate()} assertions using built-in predicates for contact, clash-freedom, connectivity, and ground support. A shared manual documents the API and workflow guidance (Supp.\ \cref{sec:supp_task_prompts}).

Adapters decouple harness and backend, instantiated here with Rhino~8~\cite{mcneel2023rhino8} across native harnesses (\cref{sec:setup}). Each run starts in an isolated workspace with task inputs, prompt, manual, Python, and modeling document, with web tools, external skills, MCP, and outside retrieval disabled (Supp.\ \cref{sec:supp_prompts}). Evaluation labels, references, frozen protocols, and the offline checker (\cref{sec:eval}) remain withheld.

\subsection{Reconstruction strategies}
\label{sec:strategies}
\paragraph{Single agent.}
A single agent builds the model iteratively (\cref{fig:method}b). At step $t$, it runs $\Pi_t$ to obtain $M_t=\mathcal E(\Pi_t)$, inspects the feedback and views, measures the scan or scan-to-$M_t$ distances in Python (slicing, plane fitting, clustering), and writes $\Pi_{t+1}$. No step is scripted, so the agent freely interleaves measurement, modeling, and revision before exporting the final program, model, and report at completion or timeout.

\begin{table*}[t]
  \centering
  \scriptsize
  \setlength{\tabcolsep}{3.5pt}
  \newcommand{\tmGap}{\hspace{7pt}}
  \newcommand{\tmMiou}{\multicolumn{1}{c}{\makebox[0pt]{mIoU$\uparrow$}}}
  \begin{tabular}{@{}l@{\hspace{4pt}} rrrr@{\tmGap} rrr@{\tmGap} rrr@{\tmGap} rrr@{\tmGap} rrr@{\tmGap} rrr@{}}
    \toprule
    & \multicolumn{4}{c@{\tmGap}}{All scenes (128)}
      & \multicolumn{3}{c@{\tmGap}}{\textcolor[HTML]{4C78A8}{Interiors (73)}}
      & \multicolumn{3}{c@{\tmGap}}{\textcolor[HTML]{28796F}{Bridges (34)}}
      & \multicolumn{3}{c@{\tmGap}}{\textcolor[HTML]{9A6B16}{Tunnels (10)}}
      & \multicolumn{3}{c@{\tmGap}}{\textcolor[HTML]{A85548}{Facades (7)}}
      & \multicolumn{3}{c@{}}{\textcolor[HTML]{736494}{Heritage (4)}} \\
    \cmidrule(r{7pt}){2-5}\cmidrule(r{7pt}){6-8}\cmidrule(r{7pt}){9-11}\cmidrule(r{7pt}){12-14}\cmidrule(r{7pt}){15-17}\cmidrule{18-20}
    Method & R5$\uparrow$ & \tmMiou & $F_1^{\mathrm{surf}}\!\uparrow$ & ECR$\uparrow$
      & R5$\uparrow$ & \tmMiou & ECR$\uparrow$ & R5$\uparrow$ & \tmMiou & ECR$\uparrow$ & R5$\uparrow$ & \tmMiou & ECR$\uparrow$ & R5$\uparrow$ & \tmMiou & ECR$\uparrow$ & R5$\uparrow$ & \tmMiou & ECR$\uparrow$ \\
    \midrule
    Cloud2BIM$^{*}$ \harness{baseline} & -- & -- & -- & -- & 55.1 & 32.7 & 6.8 & -- & -- & -- & -- & -- & -- & -- & -- & -- & -- & -- & -- \\
    \midrule
    GPT-6.1 Sol \harness{Codex} & \textbf{95.6} & \textbf{63.2} & \textbf{71.9} & 27.3 & \textbf{98.5} & \textbf{59.4} & 24.7 & \textbf{94.0} & \textbf{74.0} & \textbf{29.4} & \textbf{99.1} & \textbf{51.0} & 60.0 & \textbf{70.6} & 70.0 & 14.3 & \textbf{91.0} & 58.5 & 0.0 \\
    Claude Opus 5.5 \harness{Claude Code} & 92.2 & 60.4 & 71.2 & \textbf{37.5} & 95.5 & 57.2 & \textbf{35.6} & 91.8 & 69.2 & \textbf{29.4} & 90.5 & 44.2 & \textbf{70.0} & 66.1 & \textbf{70.5} & \textbf{71.4} & 85.6 & \textbf{65.4} & 0.0 \\
    Kimi K3.1 Pro \harness{Pi} & 74.1 & 44.5 & 64.3 & 25.8 & 82.0 & 41.9 & 23.3 & 65.9 & 57.0 & 23.5 & 80.1 & 29.0 & 50.0 & 40.4 & 37.4 & 42.9 & 42.7 & 36.7 & 0.0 \\
    GPT-6.1 Terra \harness{Codex} & 72.7 & 40.2 & 63.2 & 10.9 & 79.4 & 36.7 & 6.8 & 63.5 & 51.1 & 5.9 & 87.3 & 41.0 & \textbf{70.0} & 38.9 & 23.8 & 0.0 & 51.5 & 38.5 & 0.0 \\
    Qwen 3.8 Max \harness{Pi} & 38.5 & 19.5 & 54.7 & 11.7 & 59.0 & 27.8 & 12.3 & 8.6 & 8.2 & 5.9 & 15.1 & 3.7 & 10.0 & 22.0 & 20.6 & 42.9 & 5.0 & 2.2 & 0.0 \\
    GPT-6.1 Terra (team) \harness{Codex} & 89.8 & 58.4 & 69.3 & 21.1 & 92.5 & 56.2 & 17.8 & 88.6 & 66.0 & 8.8 & 98.4 & 48.0 & \textbf{70.0} & 61.6 & 60.1 & 57.1 & 78.4 & 56.2 & 0.0 \\
    \bottomrule
  \end{tabular}
  \caption{\textbf{Main results} across all \nscenes\ scenes. Small text after each agent gives its harness. R5 and mIoU are scene means (\%), scoring failed runs as zero, while $F_1^{\mathrm{surf}}$ averages valid submissions. Best results among the six agent rows are bolded (omitted when all tie at zero). Bootstrap intervals and paired significance tests are in Supp.\ \cref{tab:ci}. Terra (team) uses the parallel strategy (\cref{sec:strategies}). $^{*}$Cloud2BIM~\cite{zbirovsky2025cloud2bim} is a heuristic interior baseline modeling walls, slabs, doors, and windows, with mIoU over those classes (Supp.\ \cref{sec:supp_baseline}).}
  \label{tab:main}
\end{table*}

\paragraph{Parallel asynchronous agents.}
Large software projects rely on teams, and a large scene may likewise exceed what one agent can model in time. We therefore propose parallel asynchronous reconstruction, adapting two practices from software engineering (\cref{fig:method}c). First, a lead delegates tasks to up to three workers and continues modeling while they run. Second, workers never edit the shared model directly, instead working on private copies and submitting proposals for the lead to review and merge. Because the shared model evolves concurrently, tasks and proposals carry versions, and stale proposals are rejected and can be reissued. Unlike code merging, locally built components must also meet cleanly at shared physical interfaces such as wall--slab joints, which the lead inspects upon integration. Unlike staged workflows~\cite{pan2026multiagent} with fixed steps and roles, the lead freely decides whether, what, and when to delegate under one shared deadline and team budget (Supp.\ \cref{sec:supp_coordination}).

\section{Experiments}
\label{sec:exp}

\subsection{Experimental setup}
\label{sec:setup}
We evaluate five LLMs across three coding harnesses: GPT-6.1 Sol~\cite{openai2026gpt61sol} and the smaller GPT-6.1 Terra~\cite{openai2026gpt61terra} with Codex~\cite{openai2025codex}, Claude Opus~5.5~\cite{anthropic2026opus55} with Claude Code~\cite{anthropic2025claudecode}, and Kimi~K3.1 Pro~\cite{moonshot2026kimik31pro} and Qwen~3.8 Max~\cite{qwen2026qwen38max} with Pi~\cite{zechner2025pi}. Each agent runs once per scene at high reasoning effort under a 45-minute budget in Rhino~8 using standard coding tools (shell, file editing, and image reading). Every agent receives the same task prompt and seven-step verification guidance (Supp.\ \cref{sec:supp_task_prompts}), inspecting its work solely via Python measurements, rendered views, and in-script \texttt{validate()} assertions without access to our offline model checker. We also evaluate Terra as a team (\cref{sec:strategies}) on all \nscenes\ scenes and compare against the open-source heuristic pipeline Cloud2BIM~\cite{zbirovsky2025cloud2bim} on interiors (Supp.\ \cref{sec:supp_baseline}). On the 22-scene \texttt{core22} subset, we test cross-harness and run-to-run stability alongside three controlled interventions on Sol: providing ground-truth point labels, adding explicit engineering rules to the prompt, and a two-step ablation removing first workflow guidance from the manual and then verification guidance (Supp.\ \cref{sec:supp_task_prompts,sec:supp_matrix,sec:supp_interventions}).

\subsection{Main results}
\label{sec:main_results}

\begin{figure*}[t]
  \centering
  \includegraphics[width=\linewidth]{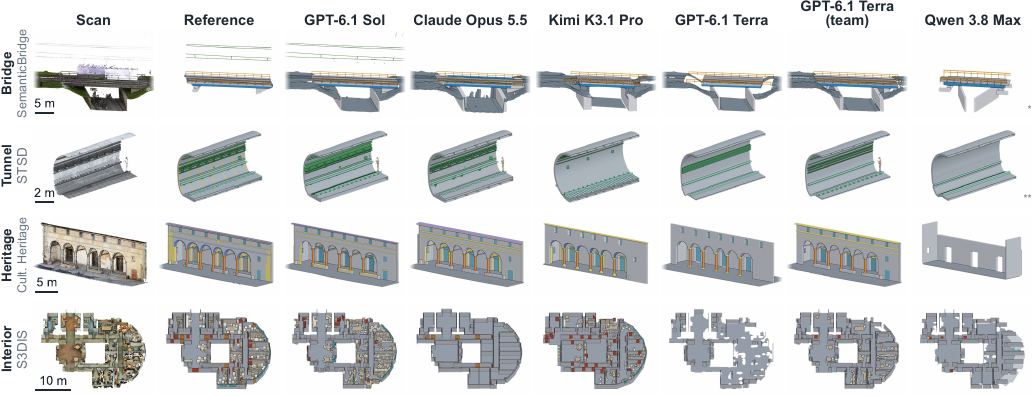}
  \caption{\textbf{Reconstructed models compared with the input scan and reference.} Colors denote component semantic classes, and tunnel and interior views are cut open for visibility. * and ** mark two Qwen outputs whose raw-coordinate 5\,cm scan coverage (R5) is 0\% due to frame/unit errors and are transformed for display only: (*) rotated $180^\circ$ to undo a flipped local frame, and (**) scaled by 1,000 from metres to millimetres. Additional comparisons across 48 scenes appear in Supp.\ \cref{fig:gallery_interior_a,fig:gallery_interior_b,fig:gallery_bridge,fig:gallery_mixed}.}
  \label{fig:qualitative}
\end{figure*}

\paragraph{Overall performance.}
High scan coverage often conceals missing components, wrong semantic classes, and broken construction relations (\cref{tab:main}, \cref{fig:teaser,fig:qualitative}). Sol and Opus cover 95.6\% and 92.2\% of scan points within 5\,cm, while the other three single agents stay below 75\% and Qwen fails to deliver a valid model on 55 scenes. Although 44.5\% of the 640 single-agent runs exceed 90\% R5, their median reference surface $F_1$ reaches only 72.3\% even at a 10\,cm tolerance, and precision exceeds recall in 74.1\% of cases due to omitted geometry (Supp.\ \cref{tab:reference-consistency}). Performance drops sharply on semantics and engineering relations, peaking at 63.2\% mIoU and 37.5\% full-scene ECR.

\paragraph{Ranking across axes.}
Fitting the scan most closely does not produce the most valid engineering model. Sol leads on scan coverage ($p<0.001$) and mIoU ($p=0.009$) and ties Opus on reference surface $F_1$ ($p=0.54$). In contrast, Opus leads on engineering conformance by 10.2\,pp ($p=0.041$), scoring highest on all nine rules (\cref{fig:rule_conformance}) and retaining its lead when resource-excluded or failed runs are removed (Supp.\ \cref{tab:ci,tab:aggregation}).

\paragraph{Domain differences.}
Performance reflects the structural and spatial demands of each domain. Coverage is near complete in interiors and tunnels, and four agents reach 50--70\% ECR in tunnels where continuous linings and track beds involve no multi-member column or beam junctions. On interiors, the heuristic baseline Cloud2BIM trails every agent in coverage (55.1\% R5) because its storey-height assumptions miss walls in 18 of 73 scenes. Street-scale facades (52--189\,m) scanned from single-sided vehicle trajectories limit single-agent coverage to at most 70.6\% under the 45-minute budget, whereas parallel collaboration lifts Terra from 38.9\% to 61.6\% R5 and from 0\% to 57.1\% ECR. Arcaded heritage buildings are the most demanding: curved vaults, columns, and dense masonry interfaces require simultaneous non-planar fitting and tight multi-element junctions, leaving localized clashes or obstructed openings in every agent model (0\% ECR).

\paragraph{Generalization and stability.}
Agent rankings transfer closely to the six held-out proprietary bridges, where Sol reaches 94.4\% R5 versus 93.9\% on public bridges (per-dataset results in Supp.\ \cref{tab:datasets}). Results are also stable across harnesses and repeated \texttt{core22} runs (Supp.\ \cref{tab:repeats}): switching Sol and Opus to the shared Pi harness shifts R5 and mIoU by under 3\,pp, and Opus consistently leads Sol in ECR across every run (22.7--27.3\% vs.\ 9.1--18.2\%).

\begin{figure}[t]
  \centering
  \includegraphics[width=0.85\linewidth]{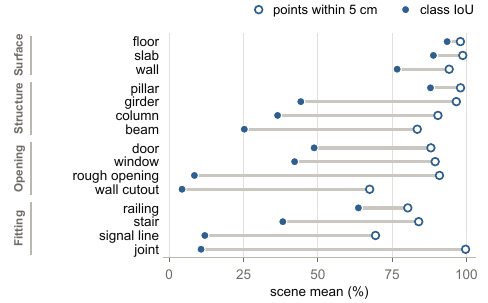}
  \caption{\textbf{Coverage and IoU per class} (Sol and Opus mean). Bars span from class IoU to 5\,cm coverage (per-agent values in Supp.\ \cref{tab:class}).}
  \label{fig:class_gap}
\end{figure}

\subsection{Analysis}
\label{sec:analysis}

\begin{figure*}[t]
  \centering
  \includegraphics[width=\linewidth]{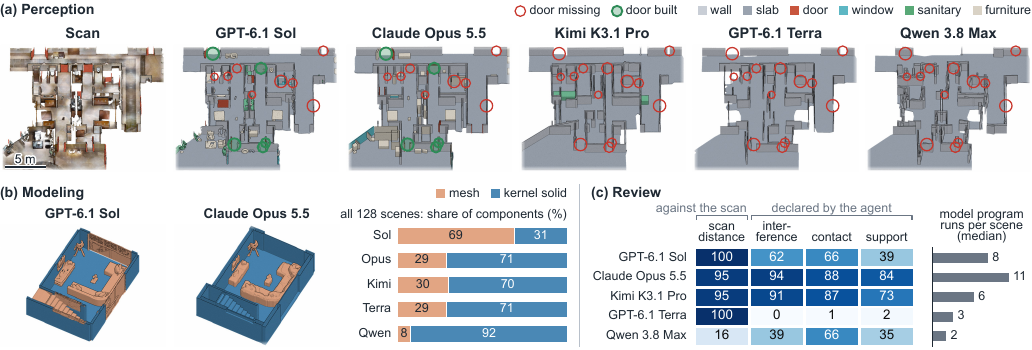}
  \caption{\textbf{What agents do, and where errors enter.} (a) An interior region in the scan and five reconstructed models, with rings marking scan door locations. (b) Sample outputs by Sol and Opus alongside each agent's mean share of meshes and kernel solids across \nscenes\ scenes. (c) Share of runs (\%) with scan checks or declared interference, contact, and support checks, and median program executions.}
  \label{fig:trajectory}
\end{figure*}

\paragraph{Component recognition.}
Semantic errors arise mainly from missed coplanar or fine components rather than poor surface fitting. In \cref{fig:class_gap}, dominant surfaces (floors, slabs, walls) reach 77--94\% IoU, whereas columns, beams, stairs, doors, and windows stay below 50\%, and openings, cutouts, signal lines, and joints fall below 15\%. Yet 67--100\% of the scan points on those missed classes still lie within 5\,cm of a neighboring wall or slab, leaving almost no distance residual. Because agents parse scans via slicing, histograms, and clustering in over 80\% of runs, they detect geometric steps while merging flush elements into host surfaces. In one 38-door interior, Kimi, Terra, and Qwen build zero doors while Sol and Opus build only 13 and 16 (\cref{fig:trajectory}a). Supplying ground-truth point labels to Sol on \texttt{core22} lifts mIoU by 27.1\,pp to 88.6\% ($+2.2$\,pp R5) but does not improve engineering conformance (\cref{tab:interv}), showing that recognizing components is necessary yet insufficient for valid assembly.

\begin{figure}[t]
  \centering
  \includegraphics[width=0.85\linewidth]{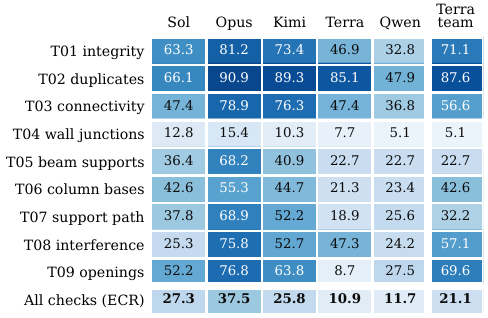}
  \caption{\textbf{Rule conformance.} Cells report RCR (\%), the share of applicable scenes passing each rule completely. The bottom row reports ECR.}
  \label{fig:rule_conformance}
\end{figure}

\begin{table}[t]
  \centering
  \footnotesize
  \setlength{\tabcolsep}{3.0pt}
  \begin{tabular}{@{}lrrrr@{}}
    \toprule
    Condition & R5$\uparrow$ & mIoU$\uparrow$ & ECR$\uparrow$ & Mean $\overline{q}_r\!\uparrow$ \\
    \midrule
    Default baseline (Sol, \cref{tab:main}) & 94.5 & 61.9 & 18.2 & 58.2 \\
    \midrule
    \multicolumn{5}{@{}l}{\textit{Input addition (prompts unchanged)}} \\
    + Ground-truth point labels & 96.5 & 88.6 & 9.1 & 54.2 \\
    \midrule
    \multicolumn{5}{@{}l}{\textit{Prompt addition (inputs unchanged)}} \\
    + Explicit engineering requirements & 93.6 & 62.6 & 18.2 & 53.9 \\
    \midrule
    \multicolumn{5}{@{}l}{\textit{Chained ablation (inputs unchanged)}} \\
    $-$ Workflow guidance (tools-only) & 94.9 & 61.7 & 4.5 & 65.9 \\
    \quad $-$ Verification guidance & 90.3 & 60.2 & 18.2 & 64.1 \\
    \bottomrule
  \end{tabular}
  \caption{\textbf{Controlled interventions on Sol (\texttt{core22}).} All arms modify the default Sol configuration of \cref{tab:main} (45-minute budget, full tool manual, verification guidance, and scan-only input), reporting metrics in \%. The prompt additions and chained ablations (first removing workflow guidance from the tool manual to leave only the tool API reference, then removing the verification guidance) are detailed in Supp.\ \cref{sec:supp_task_prompts,sec:supp_interventions}. R5 and mIoU average valid submissions (excluding one failed run under label assistance), whereas ECR and the nine-rule mean unit pass rate $\overline{q}_r$ (\cref{sec:eval}) cover all 22 scenes as in Supp.\ \cref{tab:repeats}, where baseline ECR varies over 9.1--18.2\% and $\overline{q}_r$ over 58.2--70.2\% across repeated runs.}
  \label{tab:interv}
\end{table}

\paragraph{Engineering rules and modeling.}
Chasing tighter scan fit hurts engineering consistency when agents approximate noisy surfaces with fragmented meshes instead of clean solids (\cref{fig:rule_conformance}). Although both Sol and Opus model unexplained point clusters, triangle meshes make up 69\% of Sol's components versus roughly 30\% for Opus, Kimi, and Terra (\cref{fig:trajectory}b). Sol splits surfaces into dense meshes that follow local scan irregularities, producing 861 components on one interior and $1.6\times$ the reference wall count across scenes ($0.9\times$ for Opus). This fragmentation causes material clashes and broken load paths: Sol trails Opus on interference (RCR-T08: 25.3\% vs.\ 75.8\%) and support paths (RCR-T07: 37.8\% vs.\ 68.9\%), while wall junctions (T04) remain the hardest rule for every agent ($\leq$15.4\% RCR). Adding explicit prompt rules against fragmentation, clashes, open joints, and overfitting fails to improve Sol's conformance (\cref{tab:interv}, 18.2\% ECR and $\overline{q}_r$ from 58.2\% to 53.9\%), showing that verbal instructions alone do not yield reliable 3D reconstruction.

\begin{figure}[t]
  \centering
  \includegraphics[width=\linewidth]{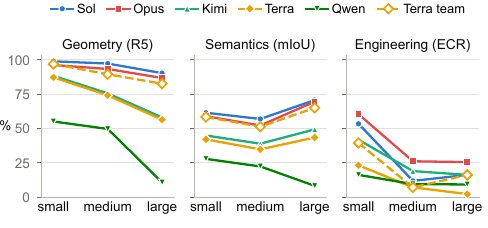}
  \caption{\textbf{The three axes by scene extent} across all \nscenes\ scenes, binned by bounding-box diagonal: small ($\leq$24\,m, 43), medium (24--51\,m, 42), and large ($>$51\,m, 43). Dashed lines show the Terra team (sensitivity in Supp.\ \cref{sec:supp_scale}).}
  \label{fig:scale_collab}
\end{figure}

\paragraph{Scene scale.}
Larger scenes widen performance gaps across agents (\cref{fig:scale_collab}). From small ($\leq$24\,m) to large ($>$51\,m) scenes, coverage drops by about 9\,pp for Sol and Opus but by roughly 31\,pp for Kimi and Terra, while ECR falls from 60.5\% to 25.6\% for Opus and from 53.5\% to 16.3\% for Sol. Pooled mIoU appears higher in the large bin only because 33 of its 43 scenes are bridges and facades with smaller class vocabularies, whereas within interiors mIoU declines steadily with scene extent (Spearman $\rho$ from $-0.42$ to $-0.32$ across binnings, Supp.\ \cref{sec:supp_scale}). Room-scale benchmarks therefore understate capability gaps on full structures.

\paragraph{Multi-agent collaboration.}
Parallel collaboration relieves context and time limits when a single agent cannot cover a large scene, but adds interface coordination overhead once coverage is already high (\cref{tab:main,fig:scale_collab}). Running Terra as a team improves every axis across all \nscenes\ scenes: R5 rises from 72.7\% to 89.8\%, mIoU from 40.2\% to 58.4\%, $F_1^{\mathrm{surf}}$ from 63.2\% to 69.3\%, and ECR from 10.9\% to 21.1\% (\$2.79 vs.\ \$0.17 per scene), with coverage gains growing from 9.7\,pp on small scenes to 26.2\,pp on large ones. In contrast, running Sol as a team on \texttt{core22} brings no net benefit ($-4.1$\,pp R5 and 18.2\% to 13.6\% ECR at $2.7\times$ cost, Supp.\ \cref{tab:efficiency,tab:collab,sec:supp_collab}), because partitioning already well-covered structures across workers introduces boundary mismatches on bridges without resolving semantic or relational errors.

\subsection{What agents verify}
\label{sec:error_results}
Inspecting all 640 single-agent script histories and 35 sampled trajectories across seven scenes (Supp.\ \cref{tab:trajectory_counts,sec:supp_process}) reveals why geometry outpaces semantics and relations: the point cloud directly penalizes surface misfits, whereas wrong classes and broken connections produce no distance alarm.

\paragraph{Scan residuals close the loop on geometry.}
Missing or shifted surfaces leave clear point-to-model gaps. Over 95\% of runs for Sol, Opus, Kimi, and Terra measure scan-to-model distances (\cref{fig:trajectory}c), and stronger agents revise their programs more often (median 11 executions for Opus and 8 for Sol). Our chained ablation on \texttt{core22} (\cref{tab:interv}) isolates what drives this loop: removing workflow guidance from the manual leaves R5 (94.9\% vs.\ 94.5\%) and mIoU (61.7\% vs.\ 61.9\%) unchanged, with ECR and $\overline{q}_r$ within repeat variance (Supp.\ \cref{tab:repeats}). Removing the verification prompt next drops R5 by 4.6\,pp (16 scenes down, one up), while mIoU ($-1.5$\,pp) and $\overline{q}_r$ (64.1\%) barely move. Capable agents can plan modeling steps on their own, and explicit verification prompts mainly drive geometric repair where scan residuals exist without fixing silent semantic or relational errors.

\paragraph{Self-verification is bounded by agent knowledge.}
Although our environment provides geometric \texttt{validate()} predicates and full Python access, raw scans carry no oracle specifying which component classes or engineering connections must exist. Catching a closed door merged into a wall or an open structural joint requires agent knowledge to combine geometric, visual, and domain cues into independent checks. Yet across the 35 sampled trajectories, agents almost never recheck semantic classes against rendered views or scan intensity after building components, aside from one elevation check by Opus in \texttt{ur6} (Supp.\ \cref{fig:traj_opus}). Similarly, while Opus (84--94\%) and Kimi (73--91\%) frequently write \texttt{validate()} assertions for interference, contact, and support (\cref{fig:trajectory}c), these assertions mostly test relations hardcoded in their own build scripts, sometimes self-referentially (such as comparing counts against generating lists or thicknesses against model constants, Supp.\ \cref{sec:supp_process,fig:traj_kimi}). Omitted openings and mislabeled solids remain internally consistent and pass unnoticed.

\section{Discussion and Limitations}
\label{sec:discussion}
Our evaluation points to three takeaways for agentic Scan-to-BIM. \textbf{(1)~Raw scans penalize surface misfits, not missing classes or broken joints}. Point-to-model distance gives a reliable geometric check, though chasing it with dense meshes hurts solid quality, and raw scans offer no validator~\cite{yang2026dreamhouse} for required components or connections (our checker is offline and reference-calibrated). Adding a general geometric linter at runtime can flag mesh defects or unasserted clashes, yet still misses flush doors merged into walls, wrong semantic classes, or open corridors versus missing joints, which pass vacuously on $M$ unless checked against the scan. Self-verification is thus bounded by what an agent knows to question, requiring it to combine geometric, visual, and engineering cues into independent hypotheses rather than grading its own build assumptions. \textbf{(2)~Semantic labels and explicit rules do not ensure valid 3D assembly}, as neither ground-truth labels ($+27.1$\,pp mIoU), auxiliary drawings and photos (Supp.\ \cref{tab:private_evidence}), nor explicit engineering prompts (\cref{tab:interv}) raise conformance. \textbf{(3)~Parallel teams expand large-scene coverage at the cost of boundary mismatches}, relieving single-agent budget limits while adding coordination overhead where sub-models meet. Improving Scan-to-BIM agents will require stronger structural and solid-modeling priors during training, paired at runtime with multi-view and learned strategy to verify construction relations (openings, classes, and shared joints) against the scan.

\paragraph{Limitations.}
Our benchmark takes registered scans and class vocabularies (not raw images or unregistered scans) under a 45-minute Rhino~8 budget, with uneven domain sizes (4 heritage; 7 facades with LoD3 shells) and Sol ablations on \texttt{core22}. We omit learned and timed human baselines because no unified method spans all five domains. The $+27.1$\,pp mIoU gain favors pairing a 3D segmentation network with agent, though point labels alone give little geometric gain and do not raise engineering conformance.

\section{Conclusion}
\label{sec:conclusion}
\bench benchmarks autonomous Scan-to-BIM across 128 scenes and five domains with three-axis evaluation and asynchronous collaboration. Across 1,028 reconstructions, agents close the loop on scan residuals to reach 95.6\% coverage yet peak at 63.2\% mIoU and 37.5\% engineering conformance, as dense meshes trade structural consistency for surface proximity and parallel teams mainly relieve large-scene budget limits. Without runtime ground truth for flush elements or construction relations, neither built-in geometric predicates nor prompted rules overcome self-referential checks or 3D interface errors. Bridging surface fitting and engineering modeling requires grounding domain knowledge in geometric, visual, and structural cues to hypothesize, verify, and enforce component classes and connections.

{\small
\bibliographystyle{ieeenat_fullname}
\bibliography{main}
}

\clearpage
\appendix
\onecolumn{\centering\Large\textbf{Supplementary Material}\par\vspace{1.0em}}
\raggedbottom
\makeatletter
\setlength{\@fptop}{0pt}
\makeatother
\setcounter{topnumber}{3}
\setcounter{bottomnumber}{2}
\setcounter{totalnumber}{4}
\renewcommand{\topfraction}{0.85}
\renewcommand{\bottomfraction}{0.6}
\renewcommand{\textfraction}{0.1}
\renewcommand{\floatpagefraction}{0.8}
\newcounter{suppfigsave}
\newcounter{supptabsave}
\newcommand{\suppneedspace}[1]{\par\penalty-100\begingroup\dimen0=#1\relax\dimen2=\pagegoal\advance\dimen2-\pagetotal
  \ifdim\dimen0>\dimen2 \ifdim\dimen2>0pt\vfil\fi\break\fi\endgroup}
\newcommand{\supptocfill}[1]{\nobreak\leaders\hbox to 0.6em{\hss.\hss}\hfill\nobreak\makebox[1.3em][r]{\pageref*{#1}}}
\newcommand{\supptocsec}[2]{\par\addvspace{3pt}\noindent
  \hyperref[#1]{\textbf{\makebox[1.7em][l]{\ref*{#1}.}#2}}\supptocfill{#1}\par}
\newcommand{\supptocsub}[2]{\par\noindent\hspace{1.7em}%
  \hyperref[#1]{\makebox[2.3em][l]{\ref*{#1}.}#2}\supptocfill{#1}\par}

\noindent
This supplementary material provides the experimental details and additional results that support the main paper. \Cref{sec:supp_prompts} describes the experimental protocol, the coordination of agent teams, the Scan-to-BIM baseline, and the task instruction and tool manual given to the agents. \Cref{sec:supp_eval} covers the reference models, the formal definitions of the scan, semantic, and surface metrics, and the model checker with its rules, scene protocols, and validation. \Cref{sec:supp_matrix} lists the runs and valid submissions and reports confidence intervals, paired tests, and the stability across repeated attempts and harnesses. \Cref{sec:supp_breakdown} breaks the main results down by class, dataset, and agreement with the references, and explains the unit pass rate and the resource limits of the checker. \Cref{sec:supp_interventions} gives the details of the intervention studies and of the auxiliary evidence on the private bridges. \Cref{sec:supp_scale} gives the details of the scale analysis and repeats it with scene size measured by the number of reference components. \Cref{sec:supp_collab} compares single agents and teams. \Cref{sec:supp_process} documents the trajectory analysis with three recorded runs, and \cref{sec:supp_gallery} shows qualitative examples on 48 further scenes.

\vspace{0.8em}
\noindent\textbf{Contents}\par\vspace{0.2em}
\noindent
\begin{minipage}[t]{0.475\textwidth}\small
\supptocsec{sec:supp_prompts}{Experimental Protocol and Prompts}
\supptocsub{sec:supp_coordination}{Team coordination records}
\supptocsub{sec:supp_baseline}{Scan-to-BIM baseline}
\supptocsub{sec:supp_task_prompts}{Prompts and tool manual}
\supptocsec{sec:supp_eval}{Reference Models and Evaluation}
\supptocsub{sec:supp_reference}{Reference models}
\supptocsub{sec:supp_scoring}{Scan and semantic metrics}
\supptocsub{sec:supp_surface}{Reference surface agreement}
\supptocsub{sec:supp_eng}{Model checker}
\supptocsec{sec:supp_matrix}{Experiment Matrix and Uncertainty}
\end{minipage}\hfill
\begin{minipage}[t]{0.475\textwidth}\small
\supptocsec{sec:supp_breakdown}{Detailed Results}
\supptocsub{sec:supp_classes}{Classes and datasets}
\supptocsub{sec:supp_ref_agreement}{Agreement with the references}
\supptocsub{sec:supp_eng_results}{Unit pass rate and resource limits}
\supptocsec{sec:supp_interventions}{Intervention Studies}
\supptocsec{sec:supp_scale}{Scale Analysis}
\supptocsec{sec:supp_collab}{Single-Agent and Team Collaboration}
\supptocsec{sec:supp_process}{Trajectory Analysis Details}
\supptocsec{sec:supp_gallery}{Qualitative Examples}
\end{minipage}
\vspace{0.6em}

\section{Experimental Protocol and Prompts}
\label{sec:supp_prompts}

\paragraph{Models and harnesses.}
We evaluate \texttt{gpt-6.1-sol} and \texttt{gpt-6.1-terra} in Codex CLI (version 0.166), \texttt{claude-opus-5-5} in Claude Code (2.1.292), and \texttt{kimi-k3.1-pro} and \texttt{qwen3.8-max-0902} in Pi (1.4.0), all at high reasoning effort. Each harness runs in its non-interactive mode (\texttt{codex exec}, \texttt{claude -p}, and \texttt{pi --mode json}). A script starts one harness process per run, passes the task instruction on standard input, sets the model and reasoning effort on the command line, and disables approval prompts so that the agent acts without confirmation. The harness comparison runs \texttt{gpt-6.1-sol} and \texttt{claude-opus-5-5} in Pi on the 22 \texttt{core22} scenes (\cref{tab:repeats}).

\paragraph{Shared reconstruction framework.}
Each run and each team worker launches in its own isolated workspace with the permitted inputs, task instruction, tool manual, Python environment (\cref{sec:supp_task_prompts}), and a fresh Rhino~8 instance. The workspace is laid out as follows.
\begin{promptbox}{Workspace layout}
\begin{verbatim}
workspace/
├── TASK.md                 # task instruction
├── AGENTS.md               # tool manual (CLAUDE.md in Claude Code)
├── input/
│   ├── cloud.laz           # point cloud
│   └── classes.json        # class specification
├── tools/
│   └── rh.py               # command-line interface to Rhino
├── templates/
│   └── model_template.py   # template of a model script
├── models/                 # generic example scripts; the agent adds its model scripts
├── work/                   # created by the agent: analysis scripts, images, exports
├── runs/                   # created per script execution: text report, rendered views
└── output/
    └── REPORT.md           # written by the agent: report
\end{verbatim}
\end{promptbox}
Web tools, user memories, external skills, plugins, and MCP connections are disabled, so agents act only through the standard coding primitives of their harness (shell execution, file editing, and image reading). They reach Rhino through the command-line interface described in the tool manual (\cref{sec:supp_task_prompts}), which executes model scripts, creates and queries class-tagged geometry, renders multi-angle views, and saves and exports the model. This interface contains no scan-processing or component-detection routines: agents measure the point cloud and the exported model with their own Python code. A run ends when the agent finishes or after 2,700\,s. Five minutes before this limit, the harness injects a one-time notice into the agent's context, stating the remaining time and that the model and report will be submitted as they are when the run is stopped, and asking the agent to finish its work. When the run ends, the harness exports the model from the Rhino document, so the agent writes no export file itself. The export consists of the native Rhino file, which keeps each component as it was modeled (kernel solid, surface, or mesh), a neutral export, and an IFC4 file generated from the neutral export with one element per component. In the neutral export, every submitted object of the Rhino document is a separate component instance with its own identifier, class tag, and boundary surface, so two walls of the same class remain two components. The boundary is stored as a triangle mesh, and kernel geometry is tessellated to within 1\,mm of its surfaces. All metrics and the model checker read the neutral export, so kernel solids and meshes are evaluated in the same way. There are no token, turn, or cost caps, and each harness keeps its native context management.

\subsection{Team coordination records}
\label{sec:supp_coordination}
This section details the parallel asynchronous team of \cref{sec:strategies} (\cref{fig:method}c), in which a lead agent delegates work to up to three workers and integrates their results into the shared model. Coordination runs through two kinds of written records: a \emph{task} is the lead's assignment to a worker, and a \emph{proposal} is the worker's reply with its results. Tasks record an identifier, revision, dependencies, evidence requirements, and deliverables. Workers receive the global task, tool manual, and current assignment, and may measure, build, or review without fixed roles. Each worker has a private workspace and Rhino~8 instance, which prevents direct concurrent edits to the lead's document. Proposals are immutable and carry their task and base revisions, candidate geometry, stable entity identities, evidence, and open issues, and a proposal built on an outdated revision is rejected. Before merging a proposal, the lead checks ownership and revision currency, imports the geometry, and inspects interfaces with neighboring components because acceptance alone does not guarantee geometric compatibility. Only the lead exports the final model.

\subsection{Scan-to-BIM baseline}
\label{sec:supp_baseline}
Scan-to-BIM methods typically chain separate stages: preprocessing of the point cloud, semantic or planar segmentation, fitting of geometric primitives, reconstruction of topology such as rooms and wall connections, and creation of parametric components for export. Each method fixes these stages for one type of asset. PolyFit~\cite{nan2017polyfit} assembles a closed polygonal surface from given planar segments and RoomFormer~\cite{yue2023roomformer} predicts floor-plan polygons, so neither yields a component model. For building interiors, Ochmann et al.~\cite{ochmann2019volumetric} recover volumetric walls, slabs, and rooms, and BIMStruct3D~\cite{chamseddine2026bimstruct3d} recovers walls, doors, and columns after learned semantic segmentation. For bridges, Fang et al.~\cite{fang2025scantobimtosim} instantiate piers and decks from predefined cross-section families, and for MEP installations, Omni-Scan2BIM~\cite{wang2024omniscan2bim} recognizes components in site images from a reference mask. Among these methods, to our knowledge only Cloud2BIM~\cite{zbirovsky2025cloud2bim} releases a complete implementation that runs from a point cloud to a component model, which makes it our baseline. It targets building interiors, so we evaluate it on the interior scenes only.

Cloud2BIM uses hand-designed geometric heuristics on point coordinates alone. It detects slabs from a height histogram, walls from horizontal storey sections, and doors and windows as rectangular wall gaps. We run the released code with its released configuration on all 73 interior scenes, converting its IFC outputs by mapping walls, doors, and windows to scene-vocabulary classes and slabs to floor and ceiling classes. When room generation fails, walls and slabs are retained. The pipeline produces a valid model on 70 scenes, while the remaining three score zero. \Cref{tab:main} reports its mIoU over the classes it can produce (32.7\%). These classes hold 94\% of the scored points on average. Over all scored classes, the set used for the agents, its mIoU is 21.6\%. In 18 scenes, the height histogram yields a single horizontal surface, so no storey or wall is formed.

\subsection{Prompts and tool manual}
\label{sec:supp_task_prompts}
Each agent receives two documents. The \emph{task instruction} is passed to the harness as its prompt and stored in the workspace as \texttt{TASK.md}. The \emph{tool manual} is stored in the workspace under the file name that the harness reads as its instruction file, \texttt{CLAUDE.md} for Claude Code and \texttt{AGENTS.md} for Codex and Pi.

\paragraph{Task instruction.}
The task instruction is shown below with its own section headings. Text in brackets is our annotation. The input list is filled from the files permitted in a run. The Acceptance section holds the verification guidance, which has its own box below. The Rhino section refers to the tool manual by its file name.
\begin{promptbox}{Task instruction}
\textbf{Task}\par
Reconstruct a high-fidelity, accurate model of the structure captured in the data in \texttt{input/}, in Rhino 8.

\smallskip\textbf{Inputs}
\begin{itemize}[nosep,leftmargin=*]
\item \texttt{input/classes.json}: class specification
\item \texttt{input/cloud.laz}: point cloud (LAZ; xyz, rgb, intensity; metres, Z up)
\end{itemize}

\smallskip\textbf{Requirements}
\begin{itemize}[nosep,leftmargin=*]
\item Model every component you can identify in the data as its own object; solid elements as closed solids.
\item Terrain and vegetation may be modelled coarsely; all other components with high fidelity and accuracy.
\item Use the coordinate frame of the input: metres, Z up, no offset, rotation or scaling.
\item Give every component exactly one class from \texttt{input/classes.json}.
\item Write a short report to \texttt{output/REPORT.md}: what you modelled and how, your assumptions, what you could not model.
\end{itemize}

\smallskip\textbf{Acceptance} \emph{[verification guidance, see the box below]}

\smallskip\textbf{Rhino}
\begin{itemize}[nosep,leftmargin=*]
\item The toolkit is in this folder, its manual in \texttt{AGENTS.md}. Classes: pass \texttt{semantic="<class>"} to \texttt{m.add(...)} in a model script; objects added any other way need the object user text \texttt{scan2bim.class = <class>}.
\item Your submission is the geometry in the Rhino document when you stop: leave only the final model there. Hidden objects, objects on hidden layers, curves, points, point clouds and text dots are not submitted. \texttt{python tools/rh.py export <folder>} writes the submission as it would be exported now. Keep the final model script(s) in \texttt{models/}.
\end{itemize}

\smallskip\textbf{Rules}
\begin{itemize}[nosep,leftmargin=*]
\item Work only inside this folder and only with the files given here.
\item \texttt{python} on the PATH is Python 3.10 with numpy, scipy, pandas, open3d, laspy (LAZ), trimesh, rtree, ifcopenshell, shapely, scikit-learn, scikit-image, opencv (cv2), networkx, matplotlib, pillow, tqdm, rhino3dm (.3dm files), manifold3d (trimesh booleans) and mapbox-earcut (polygon triangulation).
\item Nobody answers questions during the run: decide yourself and record your decisions in the report.
\end{itemize}
\end{promptbox}

\paragraph{Tool manual.}
Each workspace provides a tool manual (\cref{fig:method}a) that combines two parts: Scan-to-BIM workflow guidance and the Rhino/Python tool API reference (summarized below). The agent reaches Rhino only through \texttt{tools/rh.py}, a command-line program that forwards each shell call to the running Rhino instance and returns a text report and rendered views. The model is written as a \emph{model script}, a Python file that Rhino executes with RhinoCommon, its geometry programming interface. A model script declares its units (\texttt{UNITS}) and a dictionary of measured dimensions (\texttt{PARAMS}) and defines two functions: \texttt{build} creates the components and assigns each a class, and \texttt{validate} states the agent's own checks on the result. Running a script again replaces the objects of its previous run, so the script stays the complete description of the model. The agent decides what \texttt{validate} checks. It applies the built-in predicates to the components and relations it names, for example that a beam touches the columns carrying it or that doors do not clash with walls, and it can add checks of its own on any condition it computes (\texttt{v.check}), such as a scan-to-model deviation statistic. All of these checks are evaluated on the agent's own model components and are independent of the withheld offline model checker and frozen scene protocols, which the manual does not mention. In the \emph{tools-only} ablation of \cref{tab:interv}, the Scan-to-BIM workflow guidance is removed, leaving only the tool API reference and basic script execution steps.
\begin{promptbox}{Tool manual (summary)}
\begin{itemize}[nosep,leftmargin=*]
\item \emph{Scan-to-BIM workflow guidance (omitted in tools-only):} Inspect point cloud levels, dominant plan directions, and slices, decompose components, derive robust multi-point parameters, write \texttt{build} and \texttt{validate} checks for solids, contacts, clashes, counts, and classes, run and visually inspect rendered overlays with the point cloud, iterate, and verify completion. Plus component recipes for walls, slabs, openings, structural members, and non-prismatic shapes that regularize measured outlines into design shapes rather than following every local deviation of the points.
\item \emph{Tool API reference (retained in tools-only):} Commands of \texttt{tools/rh.py} that start and query the Rhino instance (\texttt{status}, \texttt{start}, \texttt{info}), run a model script or a short code cell inside Rhino (\texttt{run}, \texttt{exec}), render views (\texttt{shot}), export the model as it would be submitted (\texttt{export}), and edit or save the document (\texttt{clear}, \texttt{undo}, \texttt{save}, \texttt{cmd}). The structure of a model script (\texttt{UNITS}, \texttt{PARAMS}, \texttt{build}, \texttt{validate}) and how to tag components with semantic classes. Checks that run automatically on every model, and predicates the agent can call in \texttt{validate} for closed solids, contact, clashes, connectivity, ground support, component counts, class tags, numeric values, and any condition the agent computes itself (\texttt{v.solid}, \texttt{v.touching}, \texttt{v.no\_clash}, \texttt{v.connected}, \texttt{v.on\_ground}, \texttt{v.count}, \texttt{v.semantic}, \texttt{v.approx}, \texttt{v.check}). Options for visual inspection, and Python examples for reading the point cloud and the exported model.
\end{itemize}
\end{promptbox}

\paragraph{Verification guidance.}
The Acceptance section of the default task instruction holds seven-step verification guidance instructing the agent to inspect the model against the scan as a surveyor or BIM engineer would before handover (summarized below).
\begin{promptbox}{Verification guidance}
\begin{enumerate}[nosep,leftmargin=*]
\item Estimate scan noise and choose an acceptance tolerance.
\item Measure scan-to-model deviation, summarize median and P95 values, inspect spatial error patterns, and correct misplaced or missized components.
\item Cluster unexplained points and decide whether each cluster requires modeling, correction, or an explanation as a scan artifact.
\item Check that visible component faces have scan support, and report inferred geometry behind occlusions, and do not add geometry merely to cover points.
\item Compare model and cloud sections at several heights and principal directions.
\item Check that component counts match the scan, dimensions are plausible, and every component has the right semantic class.
\item Record the tolerance, deviations, unexplained clusters, inferred components, and remaining errors in an acceptance report, then correct identified problems and check again.
\end{enumerate}
\end{promptbox}
These checks are stated as a minimum; agents may add any further checks the structure or data call for.
This verification guidance specifies review actions without exposing evaluator tolerances or scores. The chained ablation in \cref{sec:supp_interventions} first removes the Scan-to-BIM workflow guidance from the tool manual (leaving the \emph{tools-only} API reference) and then removes this verification guidance.

\paragraph{Explicit engineering requirements.}
This intervention tests whether stating engineering expectations in the task improves conformance. It appends the delivery criteria summarized below to the default task instruction and changes nothing else: inputs, tool manual, and budget stay the same, no modeling method or geometry format is prescribed, and the checker's rules and tolerances are not disclosed. It does not improve Sol's conformance (\cref{tab:interv}).
\begin{promptbox}{Explicit engineering requirements}
\begin{itemize}[nosep,leftmargin=*]
\item Give each identifiable engineering component an independent identity whose extent reflects the component rather than arbitrary segmentation fragments.
\item Form valid closed solids, keep thicknesses, cross-sections, openings, and connections consistent with evidence, and avoid duplicate objects, unintended intersections, disconnected joints, and implausible dimensions.
\item Preserve evidenced curvature, slopes, deformation, and necessary details rather than over-idealizing shapes or dropping required components, and record unresolved ambiguities.
\item Avoid reproducing unsupported local point fluctuations solely to bring surfaces closer to individual points, and do not relax acceptance tolerances merely to pass checks.
\end{itemize}
\end{promptbox}

\paragraph{Collaboration prompt.}
A team run differs from a single-agent run only in its coordination instructions. The lead receives the same task instruction and tool manual, followed by the delegation section excerpted below. Each worker receives the same task instruction as background, the same tool manual, and the assignment written by the lead. Excerpts are verbatim, and dots mark omitted passages, which give the command syntax of the coordination tools.
\begin{promptbox}{Collaboration prompt (excerpts)}
You may work alone or use up to three concurrent native helpers. Choose their number, task scope and timing. Any helper may perceive, measure, analyze, build, inspect or review; short questions and revised tasks are legal. No fixed role, region partition, review count or iteration count is required.

\smallskip Your \texttt{TASK.md} is the global task. Helpers receive it as \texttt{GLOBAL\_TASK.md}, plus \texttt{AGENTS.md} as their toolkit manual. Their specific scope and deliverables belong in your published assignment. [\ldots] Register a new task revision when changing scope, owner or task status; communicate that revision to the worker. [\ldots]

\smallskip For uncertain interpretations, you may choose an independent review task. [\ldots] This is optional [\ldots]

\smallskip Use stable entity IDs to track contributions selected for integration. [\ldots] Helpers publish versioned proposals with source/evidence, category when applicable, unresolved issues and candidate snapshots. Inspect them and accept selected proposals [\ldots] Stale revisions are rejected. Acceptance updates the catalog, not your live document. Integrate selected contributions yourself, [\ldots] final model consistency still requires actual validation. [\ldots]

\smallskip Validate the whole result against original observations, reconcile interfaces and unassigned work, save the final model and report in \texttt{output/}, and stop worker Rhino instances before delivery. Final integration and submission belong to you.
\end{promptbox}

\FloatBarrier
\section{Reference Models and Evaluation}
\label{sec:supp_eval}

\subsection{Reference models}
\label{sec:supp_reference}
\paragraph{Reference authoring.}
The five domain experts hold degrees in civil engineering, computational mechanics, and architecture (2--5 years of experience).

Starting models come from GPT-6.1 Sol, which attains the highest scan coverage and lowest P95 deviation among the evaluated agents (\cref{tab:main} and \cref{sec:supp_scoring}). Experts rebuild mesh approximations as discrete components, merge fragmented parts, and correct component classes and connections. Scan coverage and mIoU do not use references, and Sol's surface F1 is computed on Sol runs that did not serve as starting models. \Cref{tab:expert_comments} summarizes the experts' recurring comments on the starting models.
The model checker uses references only to calibrate its scene protocols, which constrain engineering relations and tolerances rather than reproducing reference geometry (\cref{sec:supp_eng}).

\begin{table}[H]
\centering\footnotesize
\setlength{\tabcolsep}{4pt}
\begin{tabular}{@{}lp{0.84\textwidth}@{}}
\toprule
Domain (scenes) & Representative comments \tabularnewline
\midrule
Interiors (15) & \raggedright Partitions labeled as walls. Missing doors, column guards, and bracing. Open or misaligned wall--slab joints. Disconnected or arbitrarily rerouted pipes. Bracing detached from carried elements. Ducts split into segments. Spurious geometry from scan noise and out-of-scope areas. Rippled slab meshes and false holes from scan shadows and scanner positions. Stairs, railings, slab recesses, and columns off the scan. \tabularnewline
\addlinespace
Bridges (34) & \raggedright Classes split more finely than the annotation. Terrain labeled as abutments. Missing connectors, cross beams, and railings. Off-structure objects. Gaps between abutments, piers, and deck. Unjoined deck and girder segments. Overlapping abutment walls. Abutment shapes, pier bases, pier tops, railings, and bridge ends deviating from the scan. \tabularnewline
\addlinespace
Tunnels (10) & \raggedright Missing sleepers, cable supports, and switches. Components labeled generically as attachments. Spurious cables and channels from scan shadows. Misplaced lining segment joints and ignored inclined joints. Continuous walkway supports. Cables overlapping brackets. Rails not touching the bed. Triangulated lining meshes with dents behind occlusions. Elliptical cable sections. Rails and conductor channels off the scan. \tabularnewline
\addlinespace
Heritage (4) & \raggedright Open arch--column and parapet joints. Gaps and overlaps between stairs and terrain. Gaps in cross vaults. Missing door thresholds. Inaccurate drain pipes and sculpted details. Cross vaults and domes distorted by noise, missing data, and sculpted details. \tabularnewline
\bottomrule
\end{tabular}
\caption{Representative common expert comments on the starting models of the 63 expert-authored references, by domain.}
\label{tab:expert_comments}
\end{table}

\subsection{Scan and semantic metrics}
\label{sec:supp_scoring}

\paragraph{Formal scan and semantic metrics.}
Geometric metrics evaluate only nonignored scan points $x\in X$ belonging to task classes, with coverage $R_\tau$ as in \cref{eq:geometry_main}. We report P95, the 95th percentile of $d(x,M)$, rather than RMS distance because unmodeled regions meters away from $M$ can dominate squared errors while $R_\tau$ already captures missing coverage. P95 excludes the farthest 5\% of points while still exposing large local deviations. Across valid submissions on all \nscenes\ scenes, mean P95 is 40\,mm for Sol, 246\,mm for Opus, 877\,mm for Kimi, 1,156\,mm for Terra, 2,811\,mm for Qwen, and 133\,mm for the Terra team. On the 20 \texttt{core22} scenes where both Sol arms deliver a valid model (\cref{tab:collab}), it is 44\,mm for the single agent and 234\,mm for the team. For semantic evaluation, mIoU follows \cref{eq:semantic_main}, counted over all nonignored labeled points, where $\mathcal C$ contains the scored classes that occur in the annotation or the prediction. Predicted-only task classes score zero, context points receiving a scored class count as false positives, and missing or unclassified target points count as false negatives.

Geometry is evaluated on the 1\,cm voxelized scan (the exact cloud given to agents) restricted to points whose ground-truth classes lie in the fixed task set (\cref{tab:scoring_classes}), after excluding ignored classes and annotation states. The task set is an evaluator policy rather than a per-scene class hint, and preserves dataset-native names and granularity. Most datasets are scored on all annotated classes except catch-all labels such as \emph{other} and \emph{clutter}. Where a dataset also labels furnishings, equipment, or surroundings, these classes are ignored: furniture and equipment in BIMNet, S3DIS, LivingBIM, and CRAS Labs, ground, vegetation, and traffic signs in SemanticBridge, terrain in TUM-FA\c{C}ADE, and persons in STSD. 3DSES is the exception: the building model released with the dataset contains furniture and building equipment as elements, so its Gold tier scores these classes. DeKH is scored on walls and doors, following the evaluation of its source paper~\cite{chamseddine2026bimstruct3d}. In TUM-FA\c{C}ADE, \emph{ground surface} denotes the ground inside the building envelope, whereas the ground around the building is the separate, ignored class \emph{terrain}. Nonignored points outside the task set incur false positives if assigned a scored prediction, while ignored classes and states incur neither false positives nor false negatives. Unassigned or unclassified predictions become false negatives on eligible ground-truth points. Predicted-only scored classes enter IoU with zero intersection, while absent, unpredicted classes have empty union and are omitted.

\begin{table}[H]
\centering\small
\begin{tabular}{@{}p{0.22\textwidth}p{0.75\textwidth}@{}}
\toprule Dataset / annotation tier & Scored classes \\
\midrule
3DSES Silver/Bronze & column, covering, door, exit sign, heater, lamp, railing, slab, stair, wall, window \\
3DSES Gold & column, components, covering, damper, door, exit sign, fire terminal, furniture, heater, lamp, outlet, railing, slab, stair, switch, wall, window \\
BIMNet & wall, column, beam, slab, door, window, stair, railing \\
CRAS Labs & ceiling, floor, wall, door, window, stairs \\
DeKH & wall, door \\
LivingBIM & wall, floor, ceiling, door, column, stair, railing \\
Rohbau3D & Ceiling, Slanted Ceiling, Ceiling Cutout, Floor, Wall, Drywall, Wall Cutout, Parapet, Beam, Column, Staircase, Window Cutout, Window, Door Rough Opening, Facade Element, Elevator Doors, TGA \\
S3DIS & ceiling, floor, wall, beam, column, window, door \\
UGD & wall, pillar, ceiling, floor, door, airduct, cabletray, pipe, lamp, firehydrant \\
Japan RC Bridges & abutment, girder, deck, parapet \\
SemanticBridge & abutment, superstructure, top surface, railing, pillar \\
Weilheim & abutments, railings, deck, piers, girders \\
STSD & tunnel wall, rail track, power track, walkway, pipe, cable, attachment, big bolt hole, small bolt hole, signal line, joint \\
TUM-FA\c{C}ADE & wall, window, door, roof, ground surface, outer ceiling surface \\
Cultural Heritage & arch, column, moulding, floor, door/window, wall, stair, vault, roof \\
\bottomrule
\end{tabular}
\caption{Scored task classes per dataset. Case and dataset-native spellings are preserved.}
\label{tab:scoring_classes}
\end{table}

Scan coverage uses $\tau=5$\,cm throughout. Semantic label transfer uses an on-surface tolerance of 2\,cm and nearest-assignment limit of 10\,cm for interiors, tunnels, and heritage, and 5 and 20\,cm for bridges and facades. A point first takes the class of the nearest component surface within the on-surface tolerance. Otherwise, it takes the smallest closed containing component whose boundary lies within twice the nearest-assignment limit (20 or 40\,cm), or the nearest component within the assignment limit, and remains unassigned otherwise. Unclassified components still participate in distance matching. No class-size cutoff is applied to mIoU, distances are measured in the common metric frame, and P95 averages scene quantiles by arithmetic mean rather than pooling points.

\subsection{Reference surface agreement}
\label{sec:supp_surface}
Within the fixed task region, let $\mathcal C^{\mathrm{ref}}$ be the task classes with positive reference area, and restrict both models to $\mathcal C^{\mathrm{ref}}$. Surface precision $P^{\mathrm{surf}}$ is the fraction of $M$'s surface area within $\tau=10$\,cm of $M^{\mathrm{ref}}$, and recall $R^{\mathrm{surf}}$ reverses the two models. $F_1^{\mathrm{surf}}$ combines them as in \cref{eq:geometry_main} and is zero when no area matches. Fractions are estimated via deterministic area-uniform sampling and point-to-triangle distances under identical region and class policies.

Surface F1 averages valid submissions (\cref{tab:main,tab:reference-consistency}).

\subsection{Model checker}
\label{sec:supp_eng}
The model checker returns PASS or FAIL for each applicable unit of each rule. A unit whose check cannot be completed counts as failing. One checker version and one set of frozen scene protocols evaluate every model in this paper.

\paragraph{Terms.}
A \emph{geometric predicate} is a yes-or-no test on the geometry of one or more components, such as whether a mesh is closed or whether two faces lie within a tolerance of each other. A \emph{unit} is the object on which a rule is tested once, for example a component, a wall junction, or an opening with its host. The \emph{role} of a component is the function that the protocol assigns to its class, such as wall, column, or beam. A \emph{host} is a component that carries another, such as the column under a beam or the wall around an opening, and a \emph{root} is a component at which support paths end, such as a floor slab or an abutment. The \emph{task region} is the part of the scene that the protocol evaluates. A \emph{material region} is one closed solid part of a component, and a \emph{material island} is a piece of a component's material that is not connected to the rest of it, so a wall delivered with a detached fragment has two islands. A \emph{facet group} is a component that the protocol treats as a set of mesh facets without requiring a closed or connected surface, as for door and window assemblies, finish layers, and facade surfaces. A \emph{marker} is a component that denotes an opening, recess, or joint rather than material, such as the cutouts and rough openings of Rohbau3D or the bolt holes and joints of STSD.

\paragraph{Rule inventory and scope.}
\Cref{tab:rules} states the nine rules, and \cref{tab:rule_predicates} gives their units, geometric predicates, and frozen tolerances. The rules and predicates are shared by all scenes, and tolerances follow domain and dataset defaults. The joint tolerance $\tau_{\mathrm{joint}}$ is 20\,mm (67 interior scenes) or 30\,mm (bridges, tunnels, facades, heritage, and the 6 construction-site interiors of Rohbau3D). Two components are in contact when a pair of their opposed faces lies within $\tau_{\mathrm{joint}}$ over a patch of at least $1\,\text{cm}^2$ ($10\,\text{cm}^2$ in heritage scenes) and $1\%$ of the smaller face. The checker evaluates the delivered geometry as it is: it keeps the original component identities, class tags, coordinates, and facets, and neither welds vertices beyond floating-point rounding, caps open boundaries, removes components, nor subtracts openings. The checks test geometric construction relations in the submitted representation rather than structural capacity or code compliance.

\begin{table}[H]
\centering\footnotesize
\setlength{\tabcolsep}{4pt}
\begin{tabular}{@{}lp{0.22\textwidth}p{0.65\textwidth}@{}}
\toprule
Rule & Check (unit) & Geometric predicate and frozen tolerances \\
\midrule
T01 & Representation (\emph{component}) & The protocol declares a representation for each role. Closed solids, including components made of several closed parts where the protocol permits them, need a closed 2-manifold boundary, consistent orientation, and no self-intersection, tested exactly on the delivered coordinates. Open surfaces need the same without closure. Facet groups need finite facets with positive area. \\
T02 & Duplicate (\emph{material region}) & Two regions of the same role are duplicates when their bounding boxes agree within $\epsilon$, their areas differ by at most $1\%$ and their volumes by at most $2\%$, and their surfaces lie within $\epsilon$ of each other in both directions (4,000 samples). Facet groups are duplicates when their facet sets are identical. $\epsilon$ is $2$\,mm on 80 scenes (mainly interiors, tunnels, and facades), $5$\,mm on 47 (mainly bridges), and $15$\,mm on one bridge. \\
T03 & Connectivity (\emph{component}) & A material island counts when its area is at least $\max(25\,\text{cm}^2, 1\%\text{ of the component area})$. A component may contain only one such island. In BIMNet, CRAS Labs, DeKH, and LivingBIM, a component may instead consist of several parts if they are connected by contacts or each has its own support path. \\
T04 & Wall junction (\emph{junction}) & Units are the wall junction regions declared in the protocol (32 scenes) or, in 7 scenes, the pairs of walls that lie within $0.30$\,m of each other in the submitted model. A junction passes when two walls are in contact there, directly or through one connecting wall, partition, column, or pier. Where the protocol permits it, a junction also passes without a contact patch when a vertical edge of one wall lies within $\tau_{\mathrm{joint}}$ of a vertical edge or a face of the other wall over a height of at least $0.1$\,m. A declared junction without walls fails. \\
T05 & Beam support (\emph{beam or girder}) & Each beam or girder must be in contact with at least one eligible host (\texttt{wall}, \texttt{column}, \texttt{pier}, \texttt{beam}, \texttt{girder}, \texttt{slab}, \texttt{deck}, \texttt{abutment}, etc.) on its base, side, or top faces. In BIMNet, CRAS Labs, DeKH, and LivingBIM, a beam cut by the boundary of the task region also passes. \\
T06 & Column and pier base (\emph{column or pier}) & Within its lowest $100$\,mm, each column or pier must be in contact with an eligible host on its base or side faces. Where the protocol declares a base plane (30 scenes), a base face lying on that plane within the declared tolerance also passes. \\
T07 & Support path (\emph{structural member}) & Contacts within $\tau_{\mathrm{joint}}$ between components form a directed graph. Every structural member of the scene (e.g., \texttt{wall}, \texttt{partition}, \texttt{beam}, \texttt{girder}, \texttt{column}, \texttt{pier}) must reach a declared root (e.g., \texttt{floor}, \texttt{slab}, \texttt{deck}, \texttt{abutment}, or the boundary of the task region) along this graph. \\
T08 & Interference (\emph{component}) & A component fails when it shares material with another component deeper than $\epsilon$ (as in T02) outside the permitted joints. Overlap is located on a sampling grid of $20$\,mm or coarser and confirmed on the original meshes by two independent inside tests (ray parity and winding number). Permitted joints are bounded by type, e.g., at most $0.40$\,m of penetration at wall junctions, $0.35$\,m between slab layers, and $0.60$\,m at beam--column nodes. \\
T09 & Opening clearance (\emph{opening and host}) & For each opening and each host covering more than $40\%$ of it, the share of the opening blocked by host material is measured on $20$\,mm and $10$\,mm grids. The opening fails at a blocked share of $10\%$ or more ($14\%$ in heritage scenes, $25\%$ in one interior scene), and openings narrower than $50$\,mm fail. Doors, windows, and markers that the protocol treats as hosted features must instead lie within $0.30$\,m of an eligible host ($0.03$--$0.50$\,m in 12 scenes). \\
\bottomrule
\end{tabular}
\caption{Units, geometric predicates, and frozen tolerances of T01--T09 across all \nscenes\ scene protocols. Contact means that opposed faces of two components lie within $\tau_{\mathrm{joint}}$ of each other over a finite patch.}
\label{tab:rule_predicates}
\end{table}

\paragraph{Scene protocols.}
The checker itself encodes only the shared rule predicates and knows nothing about an individual scene. A scene protocol states how the rules apply to one scene. It specifies the task region and its boundary planes, the mapping from submitted classes to component roles such as wall, column, or beam, whether each role is modeled as a closed solid, an open surface, or a facet group, the required component types, the rules in scope, and the tolerances. Most entries are domain and dataset defaults: tunnels have no columns to check, for instance, and bridge girders rest on piers. Two kinds of located requirements are specific to a scene, namely the regions where walls must join (T04) and the base planes on which columns or piers stand (T06).

Each protocol is calibrated solely on the finalized reference model of its scene. Where the reference does not pass under the default settings because these do not fit the scene, we adjust the scope of a rule, the representation of a role, or, in a few scenes, a tolerance (\cref{tab:rule_predicates}) until the reference passes every applicable rule. For example, one bridge reference has its deck slab embedded 6--12\,mm in the members below it, so its interference tolerance is 15\,mm, and one interior has windows in a slanted wall, so its blocked-share threshold is 25\%. Calibration also fixes applicability: a rule is applicable in a scene when the reference contains at least one unit of it. The protocol is then frozen and used for every model of that scene.

At scoring time, the checker reads only the submitted model and the frozen protocol and never accesses the reference, scan, or labels, so the reference enters the scores only through the protocol. Located requirements are matched to the submitted components by region and role rather than by component identity.

\paragraph{Missing components and failed runs.}
The applicable scenes $\mathcal S_r$ in \cref{eq:ecr} are fixed before agent comparison and shared by all agents regardless of omitted components (\cref{tab:engineering_collab}), and every scene has applicable rules. A unit that the protocol requires and the model does not provide, such as a declared wall junction without walls, is a failing unit. When a required component type is completely absent, every required rule whose eligible roles include that type fails in that scene, while unrelated rules remain unaffected. An applicable rule for which the model forms no unit counts as non-conforming for that rule. Failed or invalid submissions and models exceeding the resource limits (\cref{sec:supp_eng_results}) count as non-conforming on every rule.

\paragraph{Validation of the checker.}
We validate the checker in two ways: with controls built from the reference models, and with an expert review of its verdicts on agent models.

\emph{Controls.} To verify that protocols reject violations rather than passing vacuously, we test them with controls derived from the references. A \emph{preservation} control resubmits reference geometry unchanged under new component identities and must pass. A \emph{defect} control injects a minimal violation of one rule and must fail. Defects are built geometrically, for example a duplicated component, an opened wall junction, a removed support or host, or interpenetrating solids. They are placed in a region cleared of other components, so that only the targeted rule is affected, and each defect is confirmed by an independent geometric computation rather than by the checker.

\emph{Expert review.} A domain expert judged 27 units in 27 agent models without seeing the checker's verdicts. The units cover all nine rules, all five agents, and four of the five domains (15 interior, eight bridge, three tunnel, and one heritage unit). The checker passes 15 of these units and fails 12. The expert agrees with 26 of the 27 verdicts (96.3\%): with all 12 failing verdicts and with 14 of the 15 passing ones. \Cref{tab:checker_validation} lists the units by rule.

\begin{table}[H]
\centering\small
\setlength{\tabcolsep}{8pt}
\setcounter{supptabsave}{\value{table}}\setcounter{table}{18}
\begin{tabular}{@{}lccc@{}}
\toprule
Rule & Checker passes & Checker fails & Expert agrees \\
\midrule
T01 & 2 & 1 & 3 of 3 \\
T02 & 1 & 1 & 2 of 2 \\
T03 & 2 & 0 & 2 of 2 \\
T04 & 2 & 2 & 3 of 4 \\
T05 & 1 & 1 & 2 of 2 \\
T06 & 1 & 1 & 2 of 2 \\
T07 & 2 & 2 & 4 of 4 \\
T08 & 2 & 2 & 4 of 4 \\
T09 & 2 & 2 & 4 of 4 \\
\midrule
All & 15 & 12 & 26 of 27 \\
\bottomrule
\end{tabular}
\caption{Expert review of 27 unit verdicts by rule: the number of units that the checker passes and fails, and the number of units on which the expert agrees with the checker.}
\label{tab:checker_validation}
\setcounter{table}{\value{supptabsave}}
\end{table}

The one disagreement is a wall junction (T04) in a model of GPT-6.1 Sol (\cref{fig:checker_corner}). The two walls meet only edge to edge, and their faces form no contact patch. The junction passes through the second condition of T04 in \cref{tab:rule_predicates}: a vertical edge of wall A and a vertical edge of wall B lie 14.5\,mm apart, within the joint tolerance of 20\,mm, over a common height of 2.76\,m, more than the required 0.1\,m. The corner square between the two wall ends is empty, which the expert judges a defect. This condition accepts walls that meet along a line, so T04 does not test that a corner is filled. The expert confirms every sampled failing verdict, so in this sample the checker is more lenient than the expert, not stricter.

\begin{figure}[H]
  \centering
  \includegraphics[width=0.62\linewidth]{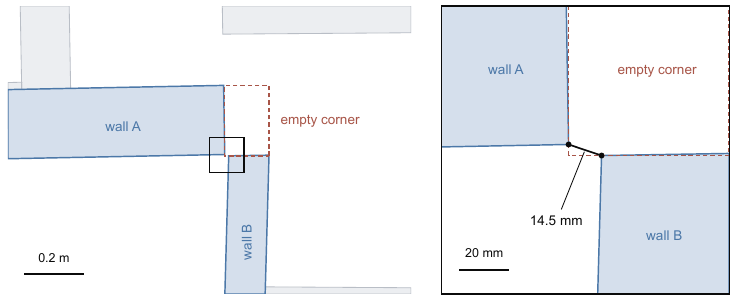}
  \setcounter{suppfigsave}{\value{figure}}\setcounter{figure}{15}
  \caption{\textbf{The wall junction on which the expert and the checker disagree} (rule T04, plan view of a model of GPT-6.1 Sol). Left: walls A and B end at the corner without overlapping, and the corner square (dashed) is empty. Other walls of the model are shown in gray. Right: the marked square, enlarged. The two dots are vertical wall edges seen from above. They lie 14.5\,mm apart, within the joint tolerance of 20\,mm, over 2.76\,m of height. T04 accepts such a line junction (\cref{tab:rule_predicates}), so the checker passes, while the expert judges the empty corner a defect.}
  \label{fig:checker_corner}
  \setcounter{figure}{\value{suppfigsave}}
\end{figure}

\FloatBarrier
\section{Experiment Matrix and Uncertainty}
\label{sec:supp_matrix}

\paragraph{Runs and valid submissions.}
The main comparison (\cref{tab:main}) comprises $5\times\nscenes=640$ single-agent runs and \nscenes\ Terra team runs. All team runs and 585 of the single-agent runs produce a scoreable model. All 55 failed runs are Qwen runs, on 49 public scenes and all six private bridges. Most of them run out of time: Qwen is often still analyzing the point cloud when the 45-minute limit arrives and has not yet executed a model program, and where it has, the programs stop with errors in the modeling code, such as RhinoCommon calls with wrong arguments or failed Boolean operations, leaving nothing to submit. In other runs, it ends on its own with an empty document. Coverage and semantic means include every scene, with zero for a failed run, and failed runs count as non-conforming in ECR and RCR. P95 averages valid submissions only (\cref{sec:supp_scoring}). \Cref{tab:efficiency} reports valid submissions and estimated API cost. The \texttt{core22} studies run Sol and Opus in the Pi harness and twice more in their native harnesses (\cref{tab:repeats}), give Sol ground-truth labels or explicit engineering requirements or remove its workflow guidance and verification guidance (\cref{sec:supp_interventions}), and run Sol as a team (\cref{sec:supp_collab}). Runs stopped by the time limit are scored whenever they export valid geometry. Including these studies and the auxiliary-evidence arms on the private bridges (\cref{tab:private_evidence}), the study scores 1,028 reconstructions.

\begin{table}[H]
  \centering
  \small
  \setlength{\tabcolsep}{3pt}
  \begin{tabular}{@{}llrr@{}}
    \toprule
    Agent & Harness & Valid$\uparrow$ & Cost (\$)$\downarrow$ \\
    \midrule
    GPT-6.1 Sol & Codex & 128 & 2.07 \\
    Claude Opus 5.5 & Claude Code & 128 & 10.37 \\
    Kimi K3.1 Pro & Pi & 128 & 2.35 \\
    GPT-6.1 Terra & Codex & 128 & 0.17 \\
    Qwen 3.8 Max & Pi & 73 & 1.90 \\
    GPT-6.1 Terra (team) & Codex & 128 & 2.79 \\
    \bottomrule
  \end{tabular}
  \caption{Valid submissions (out of \nscenes\ scenes) and mean estimated API cost per scene under the 45-minute budget. Team cost includes the lead and all workers.}
  \label{tab:efficiency}
\end{table}

\paragraph{Uncertainty and paired tests.}
\Cref{tab:ci} reports 95\% bootstrap intervals (10,000 scene resamples) for the all-scene means in \cref{tab:main}, along with paired differences between Sol and Opus and between Terra team and Terra alone. Two-sided $p$-values come from sign-flip permutation tests of the mean paired difference (100,000 permutations) for R5, mIoU, and surface F1, and from exact McNemar tests for ECR. Across individual scenes, Sol attains the higher mIoU on 54\% of scenes while Opus attains the higher surface F1 on 57\% (median difference $-2.1$\,pp), reflecting complementary per-scene behavior. Sol's mean is higher on mIoU ($p=0.009$), whereas the two agents are statistically tied on surface F1 ($p=0.54$). Opus conforms on 24 scenes where Sol fails, whereas Sol conforms on 11 where Opus fails. The Terra team conforms on 19 scenes where Terra alone fails, compared with six in the reverse direction.

\begin{table}[H]
  \centering
  \small
  \setlength{\tabcolsep}{8pt}
  \begin{tabular}{@{}lllll@{}}
    \toprule
    Agent & R5$\uparrow$ & mIoU$\uparrow$ & $F_1^{\mathrm{surf}}\!\uparrow$ & ECR$\uparrow$ \\
    \midrule
    GPT-6.1 Sol & 95.6 {\scriptsize[94.3, 96.8]} & 63.2 {\scriptsize[60.3, 65.9]} & 71.9 {\scriptsize[68.6, 75.2]} & 27.3 {\scriptsize[19.5, 35.2]} \\
    Claude Opus 5.5 & 92.2 {\scriptsize[90.3, 93.9]} & 60.4 {\scriptsize[57.2, 63.5]} & 71.2 {\scriptsize[68.3, 74.1]} & 37.5 {\scriptsize[28.9, 46.1]} \\
    Kimi K3.1 Pro & 74.1 {\scriptsize[70.7, 77.4]} & 44.5 {\scriptsize[41.7, 47.4]} & 64.3 {\scriptsize[61.2, 67.5]} & 25.8 {\scriptsize[18.0, 33.6]} \\
    GPT-6.1 Terra & 72.7 {\scriptsize[69.3, 76.0]} & 40.2 {\scriptsize[37.5, 42.9]} & 63.2 {\scriptsize[59.9, 66.4]} & 10.9 {\scriptsize[6.2, 16.4]} \\
    Qwen 3.8 Max & 38.5 {\scriptsize[31.6, 45.4]} & 19.5 {\scriptsize[15.7, 23.4]} & 54.7 {\scriptsize[49.2, 59.9]} & 11.7 {\scriptsize[6.2, 17.2]} \\
    GPT-6.1 Terra (team) & 89.8 {\scriptsize[87.8, 91.6]} & 58.4 {\scriptsize[55.3, 61.4]} & 69.3 {\scriptsize[66.0, 72.4]} & 21.1 {\scriptsize[14.1, 28.1]} \\
    \midrule
    \multicolumn{5}{@{}l}{\textit{Paired difference (pp) and two-sided $p$}} \\
    Sol $-$ Opus & $+$3.4 {\scriptsize[$+$2.1, $+$4.8]} & $+$2.8 {\scriptsize[$+$0.8, $+$4.9]} & $+$0.7 {\scriptsize[$-$1.5, $+$3.0]} & $-$10.2 {\scriptsize[$-$18.8, $-$1.6]} \\
    \quad $p$ & $<$0.001 & 0.009 & 0.54 & 0.041 \\
    Terra team $-$ Terra & $+$17.1 {\scriptsize[$+$14.9, $+$19.3]} & $+$18.1 {\scriptsize[$+$15.8, $+$20.5]} & $+$6.0 {\scriptsize[$+$4.5, $+$7.6]} & $+$10.2 {\scriptsize[$+$3.1, $+$18.0]} \\
    \quad $p$ & $<$0.001 & $<$0.001 & $<$0.001 & 0.015 \\
    \bottomrule
  \end{tabular}
  \caption{Scene-level uncertainty of \cref{tab:main} on all \nscenes\ scenes. Top: mean and 95\% bootstrap interval (\%). Bottom: paired difference in percentage points (computed before rounding) with its 95\% interval, and the $p$-value of the paired test. Surface F1 averages valid measured submissions.}
  \label{tab:ci}
\end{table}

\begin{table}[H]
  \centering
  \footnotesize
  \setlength{\tabcolsep}{3.2pt}
  \begin{tabular}{@{}lcrrrr@{}}
    \toprule
    Agent & Run & R5$\uparrow$ & mIoU$\uparrow$ & ECR$\uparrow$ & $\overline{q}_r\!\uparrow$ \\
    \midrule
    GPT-6.1 Sol & 1 & 94.5 & 61.9 & 18.2 & 58.2 \\
     & 2 & 94.9 & 61.6 & 9.1 & 70.2 \\
     & 3 & 93.6 & 60.1 & 9.1 & 58.9 \\
     & Pi & 94.2 & 61.0 & 9.1 & 70.9 \\
    \midrule
    Claude Opus 5.5 & 1 & 92.6 & 61.1 & 27.3 & 84.0 \\
     & 2 & 90.5 & 62.0 & 22.7 & 78.5 \\
     & 3 & 91.0 & 59.8 & 27.3 & 80.9 \\
     & Pi & 91.4 & 64.0 & 27.3 & 84.2 \\
    \bottomrule
  \end{tabular}
  \caption{Harness and repeated attempts on the 22 \texttt{core22} scenes. Runs 1--3 are attempts in the native harness (Codex for Sol, Claude Code for Opus), and run~1 is the run in \cref{tab:main}. \emph{Pi} runs the same agent in the Pi harness. All values are in \%.}
  \label{tab:repeats}
\end{table}

\paragraph{Repeated attempts and harness.}
\Cref{tab:repeats} tests how stable the \texttt{core22} results are. Run~1 is the default run of \cref{tab:main}, runs~2 and~3 repeat it in the native harness, and \emph{Pi} runs the same model in the shared Pi harness.

\emph{Harness.} The choice of harness has little effect. Moving from the native harness (run~1) to Pi changes R5 by $-0.3$\,pp for Sol and $-1.2$\,pp for Opus, and mIoU by $-0.9$\,pp and $+2.9$\,pp, all below 3\,pp. Engineering conformance in Pi stays close to the native attempts as well: Sol conforms on two scenes and Opus on six, with $\overline{q}_r$ of 70.9\% and 84.2\%. The differences between Sol and Opus therefore do not stem from their different native harnesses.

\emph{Repeated attempts.} Repeating a run changes coverage and semantics little. Across the three native attempts, R5 spans 1.3\,pp for Sol and 2.1\,pp for Opus, and mIoU spans 1.8\,pp and 2.2\,pp. ECR counts whole scenes, and one of the 22 scenes corresponds to 4.5\,pp: Sol conforms on two to four scenes (9.1--18.2\%) and Opus on five to six (22.7--27.3\%), with $\overline{q}_r$ of 58.2--70.2\% and 78.5--84.0\%. This variation is smaller than the gap between the two agents. On R5, ECR, and $\overline{q}_r$, their ranges do not overlap: Opus leads Sol on ECR and $\overline{q}_r$ in every run, including in Pi, and Sol leads on R5 in every run.

\FloatBarrier
\section{Detailed Results}
\label{sec:supp_breakdown}

\subsection{Classes and datasets}
\label{sec:supp_classes}
\Cref{tab:class} gives per-class coverage and IoU for Sol and Opus, and \cref{tab:datasets} lists how each dataset was acquired and breaks down the coverage and mIoU of \cref{tab:main} by dataset.

\begin{table}[H]
  \centering
  \small
  \setlength{\tabcolsep}{2.5pt}
  \def\clsh#1{\multicolumn{1}{c}{\scriptsize\shortstack{#1}}}
  \def\clshg#1{\multicolumn{1}{c@{\hspace{8pt}}}{\scriptsize\shortstack{#1}}}
  \def\z{\phantom{0}}%
  \begin{tabular}{@{}lcccc@{\hspace{8pt}}cccc@{\hspace{8pt}}cccc@{\hspace{8pt}}cccccc@{}}
    \toprule
    & \multicolumn{4}{c@{\hspace{8pt}}}{Surfaces} & \multicolumn{4}{c@{\hspace{8pt}}}{Structural} & \multicolumn{4}{c@{\hspace{8pt}}}{Openings} & \multicolumn{6}{c@{}}{Fittings, services} \\
    \cmidrule(r{8pt}){2-5}\cmidrule(r{8pt}){6-9}\cmidrule(r{8pt}){10-13}\cmidrule{14-19}
    & \clsh{floor} & \clsh{slab} & \clsh{tunnel\\wall} & \clshg{wall} & \clsh{pillar} & \clsh{girder} & \clsh{column} & \clshg{beam} & \clsh{door} & \clsh{window} & \clsh{rough\\opening} & \clshg{wall\\cutout} & \clsh{pipe} & \clsh{railing} & \clsh{stair} & \clsh{signal\\line} & \clsh{attachment} & \clsh{joint} \\
    \midrule
    \multicolumn{19}{@{}l}{\emph{R5$\uparrow$}} \\
    Sol & 98.6 & 99.4 & 99.7 & 95.3 & 97.9 & 96.1 & 96.5 & 95.9 & 92.9 & 94.0 & 83.9 & 67.6 & 91.4 & 82.8 & 93.0 & 99.0 & 92.9 & 99.5 \\
    Opus & 97.4 & 98.1 & 96.1 & 93.1 & 98.2 & 97.1 & 84.3 & 71.0 & 83.2 & 84.9 & 98.0 & 67.3 & 73.6 & 77.8 & 74.9 & 39.8 & 57.6 & 99.9 \\
    \addlinespace
    \multicolumn{19}{@{}l}{\emph{IoU$\uparrow$}} \\
    Sol & 92.9 & 89.0 & 86.6 & 76.5 & 89.0 & 54.7 & 38.7 & 16.8 & 52.9 & 45.8 & \z0.0 & \z0.0 & 85.6 & 64.6 & 35.8 & 10.4 & \z6.8 & \z3.0 \\
    Opus & 94.2 & 88.7 & 82.8 & 76.9 & 86.8 & 33.8 & 34.2 & 33.8 & 44.6 & 38.5 & 16.9 & \z8.6 & 71.0 & 62.7 & 40.6 & 13.6 & \z5.4 & 18.5 \\
    \bottomrule
  \end{tabular}
  \caption{Per-class coverage and IoU of Sol and Opus (scene means, \%), supporting \cref{fig:class_gap} (which plots the mean of the two agents). Classes keep their dataset-native names and pool the scenes of all datasets that use the name.}
  \label{tab:class}
\end{table}

\begin{table}[H]
  \centering
  \small
  \setlength{\tabcolsep}{5pt}
  \begin{tabular}{@{}lllrrrrrr@{}}
    \toprule
    Domain & Dataset & Acquisition & Scenes & Sol & Opus & Kimi & Terra & Qwen \\
    \midrule
    Interiors & 3DSES & TLS & 20 & 99.9/72.5 & 99.5/69.7 & 95.4/50.4 & 90.0/42.5 & 79.0/41.1 \\
     & BIMNet & RGB-D (Matterport) & 25 & 97.7/46.0 & 91.7/44.2 & 70.5/32.3 & 68.3/25.2 & 54.5/19.8 \\
     & CRAS Labs & TLS & 1 & 99.1/69.7 & 98.7/73.2 & 84.3/47.9 & 85.6/42.2 & 74.0/48.3 \\
     & DeKH & TLS & 4 & 94.7/67.6 & 89.5/52.3 & 74.5/48.1 & 74.4/36.4 & 42.1/20.2 \\
     & LivingBIM & TLS & 8 & 98.8/62.6 & 97.6/62.3 & 87.5/51.5 & 89.5/56.1 & 45.7/30.3 \\
     & Rohbau3D & TLS & 6 & 99.2/41.5 & 97.8/43.1 & 86.1/23.9 & 86.0/29.4 & 48.2/16.3 \\
     & S3DIS & RGB-D (Matterport) & 2 & 99.4/57.4 & 89.8/46.9 & 64.8/38.1 & 58.9/26.2 & 42.5/17.8 \\
     & UGD & TLS & 7 & 98.3/75.6 & 97.5/77.6 & 83.6/53.2 & 78.9/48.0 & 54.7/29.5 \\
    \addlinespace
    Bridges & Japan RC Bridges & LiDAR (Matterport Pro3) & 8 & 94.7/62.0 & 92.9/48.4 & 77.8/48.9 & 81.6/41.9 & 22.3/17.3 \\
     & SemanticBridge & TLS, handheld MLS & 20 & 93.6/81.5 & 95.0/78.9 & 62.5/60.7 & 58.2/57.1 & 5.8/7.0 \\
     & Weilheim & UAV photogrammetry & 6 & 94.4/65.0 & 80.0/64.8 & 61.4/55.5 & 57.0/43.0 & 0.0/0.0 \\
    \addlinespace
    Tunnels & STSD & Rail-mounted MLS & 10 & 99.1/51.0 & 90.5/44.2 & 80.1/29.0 & 87.3/41.0 & 15.1/3.7 \\
    Facades & TUM-FA\c{C}ADE & Vehicle MLS & 7 & 70.6/70.0 & 66.1/70.5 & 40.4/37.4 & 38.9/23.8 & 22.0/20.6 \\
    Heritage & Cultural Heritage & Photogrammetry & 4 & 91.0/58.5 & 85.6/65.4 & 42.7/36.7 & 51.5/38.5 & 5.0/2.2 \\
    \bottomrule
  \end{tabular}
  \caption{Acquisition and R5 / mIoU (\%) for each of the 14 datasets, including the six private Weilheim bridges. TLS denotes terrestrial laser scanning, and MLS denotes mobile laser scanning. SemanticBridge comprises 15 TLS and five handheld MLS scans. Every cell includes all scenes of the dataset, and failed runs contribute zero. Scenes receive equal weight. Vocabularies and annotation conventions differ between datasets (\cref{tab:scoring_classes}), so mIoU is comparable between agents within a row more than between rows.}
  \label{tab:datasets}
\end{table}

\subsection{Agreement with the references}
\label{sec:supp_ref_agreement}
Sol achieves the highest surface F1 on bridges, tunnels, and heritage, while Opus leads interiors at 74.3\% vs.\ 70.6\% (\cref{tab:reference-consistency}). On facades, all agents score below 26\%. These references are complete LoD3 building models that generalize facade relief, whereas the vehicle scan covers only the street side and also includes neighboring facades. Their surfaces therefore differ from the scanned surface by more than the 10\,cm tolerance over large areas, and facade surface F1 is better read as a comparison between agents than as an absolute level.

\begin{table}[H]
\centering
\footnotesize
\setlength{\tabcolsep}{0pt}
\begin{tabular*}{\linewidth}{@{\extracolsep{\fill}}l rrr rrr rrr rrr rrr@{}}
\toprule
 & \multicolumn{3}{c}{Sol} & \multicolumn{3}{c}{Opus} & \multicolumn{3}{c}{Kimi} & \multicolumn{3}{c}{Terra} & \multicolumn{3}{c}{Qwen} \\
\cmidrule(lr){2-4}\cmidrule(lr){5-7}\cmidrule(lr){8-10}\cmidrule(lr){11-13}\cmidrule(l){14-16}
Domain & $P\uparrow$ & $R\uparrow$ & $F_1^{\mathrm{s}}\!\uparrow$ & $P\uparrow$ & $R\uparrow$ & $F_1^{\mathrm{s}}\!\uparrow$ & $P\uparrow$ & $R\uparrow$ & $F_1^{\mathrm{s}}\!\uparrow$ & $P\uparrow$ & $R\uparrow$ & $F_1^{\mathrm{s}}\!\uparrow$ & $P\uparrow$ & $R\uparrow$ & $F_1^{\mathrm{s}}\!\uparrow$ \\
\midrule
All & 79.3 & 67.8 & 71.9 & 75.3 & 69.9 & 71.2 & 71.4 & 60.6 & 64.3 & 79.7 & 54.6 & 63.2 & 66.8 & 50.3 & 54.7 \\
\midrule
Interiors & 82.2 & 63.9 & 70.6 & 81.8 & 70.4 & 74.3 & 76.0 & 60.3 & 66.0 & 83.2 & 51.3 & 62.1 & 75.0 & 57.6 & 64.0 \\
Bridges & 80.4 & 77.6 & 78.5 & 71.3 & 74.2 & 71.9 & 69.3 & 62.9 & 65.1 & 79.3 & 61.0 & 67.8 & 45.2 & 36.0 & 35.7 \\
Tunnels & 94.2 & 91.4 & 92.7 & 83.1 & 80.4 & 81.5 & 85.7 & 81.5 & 83.4 & 87.2 & 81.8 & 84.4 & 73.7 & 38.0 & 42.7 \\
Facades & 23.0 & 24.9 & 23.7 & 24.7 & 27.3 & 25.5 & 21.9 & 28.6 & 23.9 & 43.2 & 19.3 & 25.0 & 13.4 & 24.2 & 17.2 \\
Heritage & 78.2 & 70.3 & 73.3 & 61.5 & 72.8 & 64.1 & 56.8 & 51.3 & 51.6 & 65.4 & 53.6 & 58.7 & 71.3 & 14.7 & 20.8 \\
\bottomrule
\end{tabular*}
\caption{Agent-to-reference agreement (\%) at 10\,cm tolerance: surface precision $P$, recall $R$, and F1 ($F_1^{\mathrm{s}}$) as equal-scene means over valid submissions. Bridges include the six private bridges.}
\label{tab:reference-consistency}
\end{table}

\subsection{Unit pass rate and resource limits}
\label{sec:supp_eng_results}
Because ECR and RCR weight scenes equally rather than units, large scenes with many components do not dominate the mean, though a single failing unit fails the scene completely. For small-cohort interventions (\cref{tab:interv}), the unit-level pass rate $\overline q_r$ captures partial conformance by averaging $q_{sr}=p_{sr}/(p_{sr}+f_{sr})$ over applicable scenes $\mathcal S_r$. Some rules test two kinds of units in the same scene, for example closed solids and facet groups in T01, or openings tested for clearance and hosted features tested for attachment in T09. The pass rate is then computed for each kind separately, and the kinds are averaged with equal weight, so that the more numerous kind does not outweigh the other. A failed run or a model beyond the resource limits enters $\overline q_r$ with a pass rate of zero on every applicable rule.

\paragraph{Resource limits and exclusions.}
The checker tests the delivered triangles exactly, so its cost grows with the size of a model. To bound this cost, every model, including the references, is subject to the same two limits: at most 2.5 million triangles per submission and 3,600\,s of checking time per model, in a single attempt. A model beyond a limit is not checked and counts as non-conforming on every rule, while its scan, semantic, and surface metrics are scored as usual. In the main comparison, this applies to 19 models, 18 of Sol and one of Terra. All of them exceed the triangle limit, with 2.6 to 12.5 million triangles. The export is not the cause. It tessellates only kernel geometry, and it keeps a mesh that the agent built itself as it is, without refining it. Of the 19 models, 13 consist of such meshes only. The other six also contain between 15 and 191 components modeled as kernel geometry, but the tessellation of these components accounts for less than 0.5\% of the triangles of each model. The triangle counts therefore come from the meshes that the agents built, not from the export. \Cref{tab:trajectory_counts} gives the shares of kernel solids and meshes for all agents. The limit also leaves room for detailed models: the references have a median of 65 thousand triangles and at most 2.2 million, and no model of Opus, Kimi, or Qwen exceeds 1.6 million.

Of the 19 excluded models, 18 are Sol's, so this convention mainly lowers Sol's ECR. To check that it does not decide the comparison between agents, \cref{tab:aggregation} recomputes ECR on two smaller sets of scenes: the 109 scenes on which no agent's model is excluded, and the 62 scenes on which the models of all five agents are checked, which also removes the failed runs. Opus leads in all three cases. Sol leads Kimi on all scenes and on the 109 scenes without exclusions, but trails it on the 62 commonly checked scenes.

\begin{table}[H]
\centering\small
\begin{tabular}{@{}lrrr@{}}
\toprule
Model & All (128) & No exclusion (109) & All checked (62) \\
\midrule
Sol & 27.3 & 32.1 & 32.3 \\
Opus & 37.5 & 43.1 & 45.2 \\
Kimi & 25.8 & 30.3 & 35.5 \\
Terra & 10.9 & 12.8 & 9.7 \\
Qwen & 11.7 & 13.8 & 24.2 \\
\bottomrule
\end{tabular}
\caption{ECR (\%) of the single-agent runs under three scene cohorts: all scenes, scenes without any resource-limit exclusion, and scenes where every agent's model is checked (no resource-excluded or failed run).}
\label{tab:aggregation}
\end{table}

\section{Intervention Studies}
\label{sec:supp_interventions}
The three interventions on \texttt{core22} each change one thing in the default Sol configuration of \cref{tab:main} and keep the 45-minute budget (\cref{tab:interv}). Label assistance adds one ground-truth class label per scan point to the inputs and leaves the prompts and the tool manual unchanged. The explicit engineering requirements are appended to the task instruction (\cref{sec:supp_task_prompts}). The chained ablation first removes the workflow guidance from the tool manual and then also removes the verification guidance (\cref{sec:supp_task_prompts}).

Label assistance yields a valid model on 21 of the 22 scenes and fails on one. All other arms produce valid models on all 22 scenes. The gains quoted in the main paper compare the label arm with the default run on these 21 scenes. There, the default run reaches 94.3\% R5 and 61.5\% mIoU, and the label arm reaches 96.5\% and 88.6\%. The gain is therefore 2.2\,pp in R5 and 27.1\,pp in mIoU. The default row of \cref{tab:interv} shows 94.5\% and 61.9\% because it averages all 22 scenes.

Removing the verification guidance lowers R5 by more than 0.5\,pp on 16 scenes and raises it on one, whereas mIoU falls on 12 scenes and rises on nine.

For engineering conformance, we compare each arm with the three runs of the unchanged default configuration (\cref{tab:repeats}). These three runs conform on four, two, and two of the 22 scenes, and their unit pass rate $\overline{q}_r$ is 58.2\%, 70.2\%, and 58.9\%.

\emph{Conforming scenes.} Two scenes conform with label assistance, four with the explicit engineering requirements, one without workflow guidance, and four without workflow and verification guidance. The arm without workflow guidance is therefore one scene below the lowest default run. The next arm of the chain, which removes the verification guidance as well, conforms on four scenes again. The three default runs already differ from each other by two scenes, so we do not attribute the single scene to the removed guidance.

\emph{Unit pass rate.} $\overline{q}_r$ is 54.2\% with label assistance, 53.9\% with the explicit requirements, 65.9\% without workflow guidance, and 64.1\% without workflow and verification guidance. The two arms without guidance lie inside the default range of 58.2\% to 70.2\%. The label and requirement arms lie about 4\,pp below its lower end.

No arm raises the number of conforming scenes or the unit pass rate above the default runs.

\paragraph{Auxiliary evidence on the private bridges.}
In an exploratory study on the six private bridge scenes, Sol is also given design drawings, site photographs, or both, with one run per scene and arm (\cref{tab:private_evidence}). The scan-only arm is the run in \cref{tab:main}. The additional evidence leaves coverage and mIoU within a few points of the scan-only arm. Engineering conformance stays low as well: no model conforms with the scan alone, with drawings, or with photographs, one of the six conforms with both, and $\overline{q}_r$ lies between 65.9\% and 70.4\% across the four arms, a smaller spread than between repeated default runs (\cref{tab:repeats}).

\begin{table}[H]
  \centering
  \small
  \setlength{\tabcolsep}{5pt}
  \begin{tabular}{@{}lrrrr@{}}
    \toprule
    Input & R5$\uparrow$ & mIoU$\uparrow$ & ECR$\uparrow$ & $\overline{q}_r\!\uparrow$ \\
    \midrule
    Scan only & 94.4 & 65.0 & 0.0 & 67.0 \\
    + design drawings & 95.0 & 64.7 & 0.0 & 65.9 \\
    + site photographs & 92.8 & 63.1 & 0.0 & 67.6 \\
    + drawings and photographs & 92.5 & 68.3 & 16.7 & 70.4 \\
    \bottomrule
  \end{tabular}
  \caption{GPT-6.1 Sol on the six private bridge scenes with auxiliary evidence (\%, one run per scene and arm). ECR counts all six scenes, and $\overline{q}_r$ averages the unit pass rate over the six rules that apply to these scenes (T01--T03 and T06--T08).}
  \label{tab:private_evidence}
\end{table}

\section{Scale Analysis}
\label{sec:supp_scale}

Scene extent is the length of the diagonal of the bounding box of the scan. The two bin boundaries, 24.3\,m and 51.0\,m, are the tertiles of the \nscenes\ scenes and are the same for all agents.

The three bins also differ in the kind of scenes they contain. The small bin holds 31 interiors, 10 tunnels, and 2 bridges. The medium bin holds 35 interiors, 6 bridges, and 1 heritage scene. The large bin holds 26 bridges, 7 interiors, 7 facades, and 3 heritage scenes. The applicable rules and the class vocabularies therefore differ between bins, and \cref{fig:scale_collab} shows the combined effect of extent, domain, vocabulary, and geometric complexity, which change together in real scans.

The difference between small and large scenes does not depend on the choice of bins. We repeated the analysis with fixed boundaries at 20 and 50\,m, with fixed boundaries at 25 and 60\,m, with a median split, with quartiles, and with quintiles. In every case and for every agent, both R5 and ECR are lower in the largest bin than in the smallest.

The decline is also visible without bins. Computed over the individual scenes, the Spearman correlation between extent and R5 lies between $-0.71$ and $-0.49$ for the five agents. Within interiors alone, it lies between $-0.60$ and $-0.42$ for R5 and between $-0.42$ and $-0.32$ for mIoU, so both also decrease with extent inside a single domain.

The team reduces the loss on large scenes. From the small to the large bin, R5 of the Terra team falls by 14.3\,pp, about half of the 30.8\,pp lost by Terra alone.

\paragraph{Scene size by reference component count.}
Scan extent measures how far a scene reaches, not how much there is to model. As a second measure of scene size, we count the components of the reference model of each scene. The count ranges from 11 to 1,388, with a median of 127. \Cref{fig:scale_components} repeats \cref{fig:scale_collab} with the scenes binned by the tertiles of this count: small (at most 75 components, 43 scenes), medium (76 to 201, 43 scenes), and large (more than 201, 42 scenes). The two measures are related but not the same. Their Spearman correlation is 0.64 over all scenes and 0.82 within interiors. The ten tunnels, for example, all lie in the small bin by extent, but eight of them lie in the large bin by component count. The small bin now holds 36 interiors, 5 bridges, and 2 facades. The medium bin holds 25 interiors, 13 bridges, 3 facades, and 2 tunnels. The large bin holds 16 bridges, 12 interiors, 8 tunnels, 4 heritage scenes, and 2 facades.

\begin{figure}[H]
  \centering
  \includegraphics[width=0.5\linewidth]{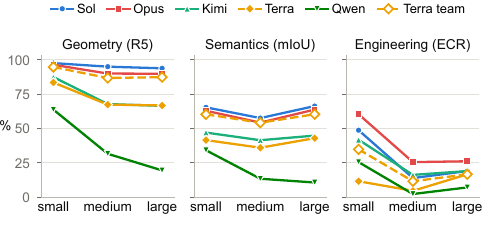}
  \setcounter{suppfigsave}{\value{figure}}\setcounter{figure}{14}
  \caption{\textbf{The three axes by reference component count} across all \nscenes\ scenes, binned by the number of components of the reference model: small ($\leq$75, 43), medium (76--201, 43), and large ($>$201, 42). Dashed lines show the Terra team. Metrics and conventions are those of \cref{fig:scale_collab}.}
  \label{fig:scale_components}
  \setcounter{figure}{\value{suppfigsave}}
\end{figure}

\emph{Coverage.} R5 is highest on the scenes with few components and falls from the small to the medium bin for every agent. From the medium to the large bin it changes by less than 1.5\,pp for every agent except Qwen. As with extent, the loss differs strongly between agents. From the small to the large bin, Sol loses 3.8\,pp and Opus 6.5\,pp, whereas Kimi loses 21.0\,pp, Terra 16.8\,pp, and Qwen 44.3\,pp. These losses are smaller than the losses by extent (8.6\,pp for Sol and 30.8\,pp for Terra), and the correlation between component count and R5 is weaker ($-0.53$ to $-0.43$ for the five agents, against $-0.71$ to $-0.49$ for extent). Coverage is therefore more closely related to how far a scene extends than to how many components it contains. The Terra team gains about 11\,pp of R5 over Terra alone in the small bin and about 21\,pp in the large bin.

\emph{Semantics.} Pooled mIoU does not decrease steadily with the component count: for every agent except Qwen it is lowest in the medium bin. The bins again differ in domain composition, so pooled mIoU mixes different class vocabularies. Within interiors, mIoU decreases with the component count for all five agents, with Spearman correlations between $-0.36$ and $-0.30$.

\emph{Engineering.} ECR falls sharply from the small to the medium bin, from 48.8\% to 14.0\% for Sol and from 60.5\% to 25.6\% for Opus, and does not fall further in the large bin (19.0\% and 26.2\%). The drop is concentrated in interiors. Opus, none of whose models is excluded by the resource limits, conforms on 21 of the 36 interiors in the small bin and on 5 of the 37 interiors with more than 75 reference components. Terra alone is the exception: its ECR is 11.6\% in the small bin and 16.7\% in the large bin. This is an effect of the tunnels. Their references have between 174 and 642 components, so eight of the ten tunnels lie in the large bin, and tunnels are the domain in which Terra conforms most often (7 of 10 scenes, against 5 of 73 interiors and 2 of 34 bridges). Five of Terra's seven conforming scenes in the large bin are tunnels, and without the tunnels its ECR in the large bin is 5.9\%, below the small bin.

These results do not depend on the choice of bins. With a median split, quartiles, or quintiles of the component count, R5 is lower in the highest bin than in the lowest for every agent, and ECR is lower for every agent except Terra. Both measures of scene size thus give the same picture: coverage and conformance are highest on small scenes, and the differences between agents grow with scene size.

\section{Single-Agent and Team Collaboration}
\label{sec:supp_collab}
Single-agent and team runs use the same scenes, protocols, checker, and evaluator, so their scores can be compared directly. The Terra comparison covers all \nscenes\ scenes (\cref{tab:main,tab:efficiency,tab:engineering_collab}), and the Sol comparison covers \texttt{core22} (\cref{tab:collab,tab:engineering_collab}).

For Terra, the team raises conformance on seven of the nine rules (\cref{tab:engineering_collab}). The largest gains are in opening clearance (T09, from 8.7\% to 69.6\%), integrity (T01, from 46.9\% to 71.1\%), and column bases (T06, from 21.3\% to 42.6\%). The mean P95 deviation falls from 1,156\,mm to 133\,mm (\cref{sec:supp_scoring}).

For Sol, the team brings no net benefit on \texttt{core22} (\cref{tab:collab}). R5 and mIoU are compared on the 20 scenes where both the single agent and the team deliver a valid model. On these scenes, R5 falls from 94.2\% to 90.1\%, and most of this difference comes from one bridge, where R5 falls by 66\,pp. mIoU changes from 60.3\% to 61.3\%. ECR counts all 22 scenes: four scenes conform for the single agent and three for the team. The team costs 2.7 times as much as the single agent.

\begin{table}[H]
\begin{minipage}[t]{0.49\textwidth}
  \centering
  \small
  \setlength{\tabcolsep}{4pt}
  \begin{tabular}[t]{@{}llrrrr@{}}
    \toprule
    Scenes & Mode & R5$\uparrow$ & mIoU$\uparrow$ & ECR$\uparrow$ & \$$\downarrow$ \\
    \midrule
    \multicolumn{6}{@{}l}{\textit{GPT-6.1 Sol, \texttt{core22}}} \\
    \texttt{core22} (20) & single & 94.2 & 60.3 & 18.2 & 2.16 \\
     & team & 90.1 & 61.3 & 13.6 & 5.81 \\
    \bottomrule
  \end{tabular}
  \caption{Single agent and team for GPT-6.1 Sol on \texttt{core22} (20 paired valid scenes for R5/mIoU and all 22 scenes for ECR; estimated USD cost includes the whole team). Terra single-vs-team comparisons appear in \cref{tab:main,tab:efficiency,tab:engineering_collab}.}
  \label{tab:collab}
\end{minipage}\hfill
\begin{minipage}[t]{0.49\textwidth}
\centering\small
\setlength{\tabcolsep}{4pt}
\begin{tabular}[t]{@{}lrrrrrr@{}}
\toprule
 & \multicolumn{3}{c}{Terra, all scenes} & \multicolumn{3}{c}{Sol, \texttt{core22}} \\
\cmidrule(lr){2-4}\cmidrule(l){5-7}
Rule & Scenes & Single & Team & Scenes & Single & Team \\
\midrule
T01 & 128 & 46.9 & 71.1 & 22 & 63.6 & 63.6 \\
T02 & 121 & 85.1 & 87.6 & 20 & 65.0 & 70.0 \\
T03 & 76 & 47.4 & 56.6 & 16 & 43.8 & 43.8 \\
T04 & 39 & 7.7 & 5.1 & 9 & 11.1 & 0.0 \\
T05 & 22 & 22.7 & 22.7 & 4 & 25.0 & 25.0 \\
T06 & 47 & 21.3 & 42.6 & 10 & 40.0 & 50.0 \\
T07 & 90 & 18.9 & 32.2 & 17 & 35.3 & 29.4 \\
T08 & 91 & 47.3 & 57.1 & 17 & 23.5 & 35.3 \\
T09 & 69 & 8.7 & 69.6 & 15 & 60.0 & 46.7 \\
\midrule
ECR & 128 & 10.9 & 21.1 & 22 & 18.2 & 13.6 \\
\bottomrule
\end{tabular}
\caption{Rule conformance (RCR, \%) of single agent and team: the share of the scenes where a rule applies in which the model passes it completely. Scene counts are shared by both arms. Resource-excluded and failed runs count as non-conforming.}
\label{tab:engineering_collab}
\end{minipage}
\end{table}

\suppneedspace{1.0in}
\section{Trajectory Analysis Details}
\label{sec:supp_process}

\paragraph{Operation counts.}
Counts cover the 640 single-agent runs in \cref{tab:main}. We search each agent's own measurement scripts and model programs (excluding shipped environment examples) after stripping comments and string literals. Each operation is defined by function and method patterns, such as nearest-point queries against the exported model, and a run counts if at least one script matches. Inspecting ten random matches per operation confirmed eight to ten true uses across all operations in \cref{tab:trajectory_counts}. Code sent directly to the kernel without being saved is not searched. Interference, contact, and support checks correspond mainly to validation calls naming components and expected relations. Geometry types come from delivered native models (boundary representations or meshes), and program executions count kernel runs of the model program.

\begin{table}[H]
  \centering
  \small
  \setlength{\tabcolsep}{5pt}
  \begin{tabular}{@{}l@{\hspace{4pt}}rrrrr@{}}
    \toprule
    & Sol & Opus & Kimi & Terra & Qwen \\
    \midrule
    \multicolumn{6}{@{}l}{\emph{Measurement (\% of runs)}} \\
    Slicing & 100 & 98 & 98 & 98 & 74 \\
    Histograms & 99 & 97 & 82 & 97 & 62 \\
    Clustering & 100 & 98 & 92 & 100 & 31 \\
    \addlinespace[1pt]
    \multicolumn{6}{@{}l}{\emph{Checks (\% of runs)}} \\
    Scan-to-model distance & 100 & 95 & 95 & 100 & 16 \\
    Interference & 62 & 94 & 91 & 0 & 39 \\
    Contact & 66 & 88 & 87 & 1 & 66 \\
    Support & 39 & 84 & 73 & 2 & 35 \\
    \addlinespace[1pt]
    \multicolumn{6}{@{}l}{\emph{Submitted geometry (\% of valid submissions)}} \\
    Kernel solids only & 3 & 44 & 31 & 54 & 84 \\
    Solids and meshes & 48 & 45 & 64 & 25 & 12 \\
    Meshes only & 48 & 11 & 5 & 21 & 4 \\
    Mean mesh share (\%) & 69 & 29 & 30 & 29 & 8 \\
    \addlinespace[1pt]
    \multicolumn{6}{@{}l}{\emph{Per run (median)}} \\
    Model program executions & 8 & 11 & 6 & 3 & 2 \\
    \bottomrule
  \end{tabular}
  \caption{Script operations and run properties across 128 runs per agent. Shares count runs with at least one matching script (runs without scripts count as zero). Valid submissions are 73 for Qwen and 128 for all others.}
  \label{tab:trajectory_counts}
\end{table}

\paragraph{Run sample.}
We sample from the public scenes on which all five single-agent runs are scored on every metric, enabling direct five-way comparisons per scan. Using a fixed random seed, we draw one scene per domain and a second scene without replacement from interiors and bridges, yielding seven scenes (35 runs): BIMNet \texttt{ur6} and 3DSES \texttt{bronze\_S142} (interiors), Japan RC Bridges \texttt{Bridge\_3} and SemanticBridge \texttt{bridge\_17\_fr\_rtc} (bridges), STSD \texttt{horse-1-6\_2} (tunnel), TUM-FA\c{C}ADE \texttt{DEBY\_LOD2\_4959322} (facade), and Cultural Heritage \texttt{5\_CB} (heritage). Because failed scenes lie outside this pool, the sample characterizes completed trajectories rather than delivery failures.

\paragraph{Reading procedure.}
Each of the 35 runs is described with the same five-part form: (i) timeline (duration, program executions, how the run ended), (ii) reading of the scan (measurements, viewed images, class decisions, decomposition, omissions), (iii) modeling (geometry per class, sources of dimensions, connections, changes between program versions), (iv) checking (checks run, their reference values, follow-up actions, checks of classes and connections, report content), and (v) the two or three largest shortfalls in the evaluation, traced to the earliest step that causes them. Every entry cites the script, program version, or step that supports it.

\begin{figure}[H]
  \centering
  \includegraphics[width=\linewidth]{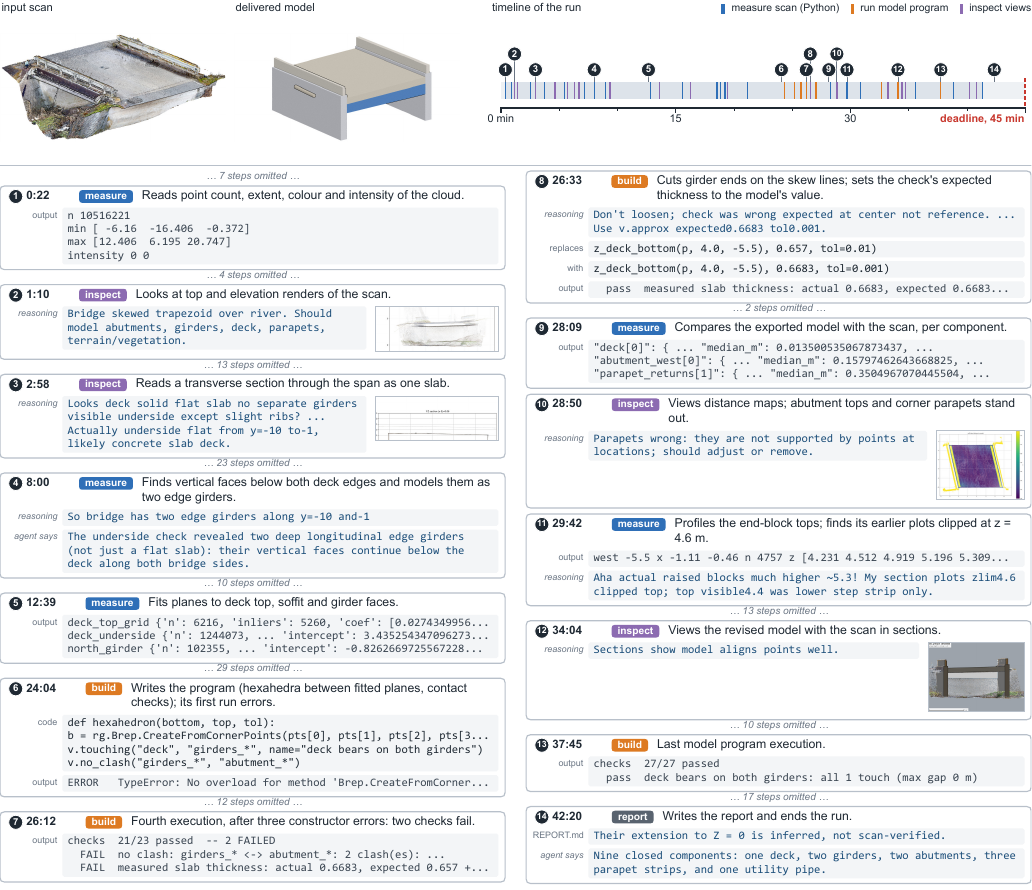}
  \caption{\textbf{Kimi K3.1 Pro on Japan RC Bridges \texttt{Bridge\_3} (bridge).} The run takes 43 minutes and executes the model program 8 times. The delivery has 9 components, lies within 5\,cm of 63.3\% of the scan points, and reaches 32.4\% mIoU. The agent first reads the span as one slab (step 3), then finds vertical faces below both deck edges and models them as edge girders (step 4). It builds the components as hexahedra between fitted planes and declares contact and interference checks (step 6). When a thickness check fails, it sets the expected value to the value of the model (steps 7--8). The comparison with the scan shows misfits at the abutment tops, and the agent finds that its earlier sections were clipped and revises the end blocks (steps 9--12).}
  \label{fig:traj_kimi}
\end{figure}

\paragraph{Cases.}
\emph{Doorway region.} A door location in \texttt{ur6} is a plan-view group of scan points labeled as door in the ground truth, counted as built if a door component lies within 0.3\,m. The region in \cref{fig:trajectory}a is selected automatically as the box containing 10--16 door locations whose Sol and Opus build shares best match their scene-wide rates.
\emph{Self-referential checks.} One Kimi script checks a written floor level against its own hardcoded constant, and several scripts compare component counts against the list that generates the components, so these checks pass by construction. In \cref{fig:traj_kimi} (steps 7--8), a failed thickness check is resolved by setting the expected value to the value of the model.
\emph{Reports.} Qwen's report on \texttt{bridge\_17\_fr\_rtc} states that systematic offsets between scan and model are below 2--3\,cm. The run did not measure scan-to-model distances: the value follows from the bin size used to fit the profiles, and the report itself notes that the comparison was not completed.

\begin{figure}[H]
  \centering
  \includegraphics[width=\linewidth]{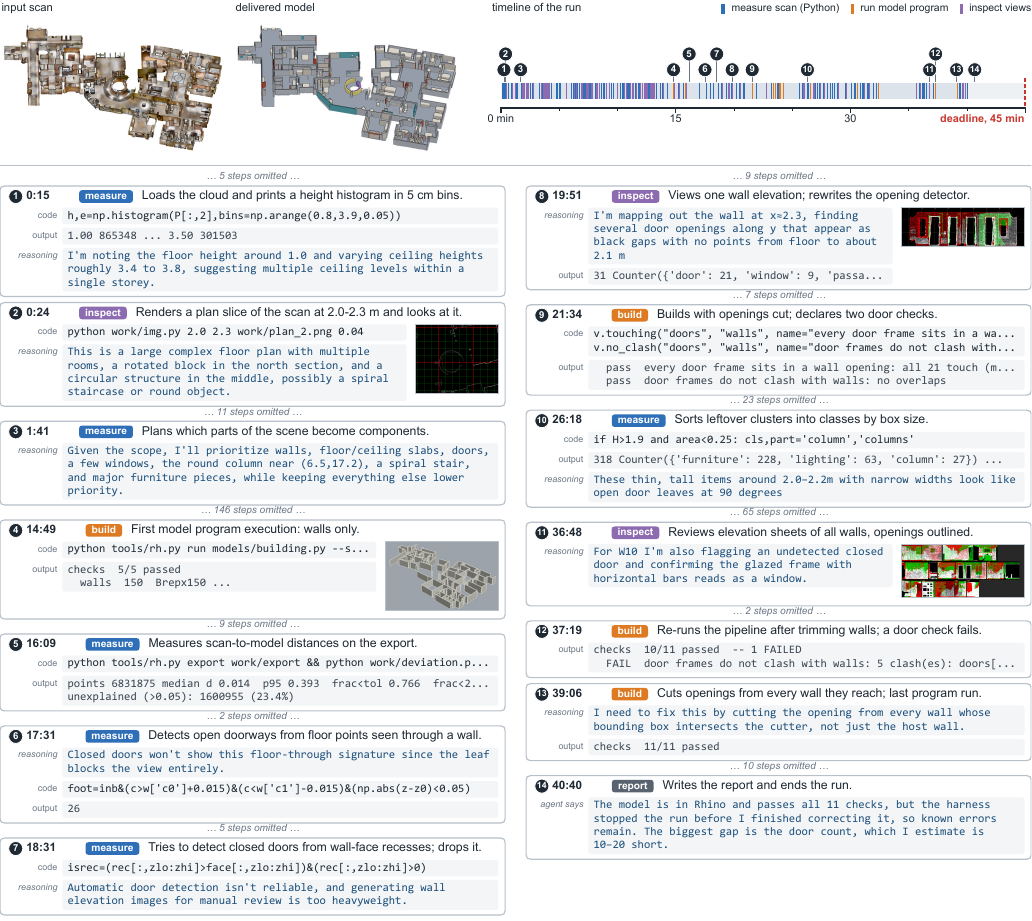}
  \caption{\textbf{Claude Opus 5.5 on BIMNet \texttt{ur6} (interior).} The run takes 41 minutes and executes the model program 14 times. The delivery has 508 components, lies within 5\,cm of 92.2\% of the scan points, and reaches 27.8\% mIoU. The agent measures the scan for 15 minutes before its first program execution and then extends the model in stages: walls, slabs, openings, and objects from clusters of unexplained points. It searches for doors as open gaps, drops a detector for closed doors (steps 6--7), and later notes an undetected closed door in an elevation sheet (step 11). Its elevation images color scan points by offset from its fitted wall plane: white within the wall, green in front, red behind, and black where there are no points. It repairs a failed interference check between doors and walls (steps 12--13) and reports that the door count is short. The run ends shortly after the five-minute notice of the harness (\cref{sec:supp_prompts}), which the agent describes as the harness stopping the run.}
  \label{fig:traj_opus}
\end{figure}

\begin{figure}[H]
  \centering
  \includegraphics[width=\linewidth]{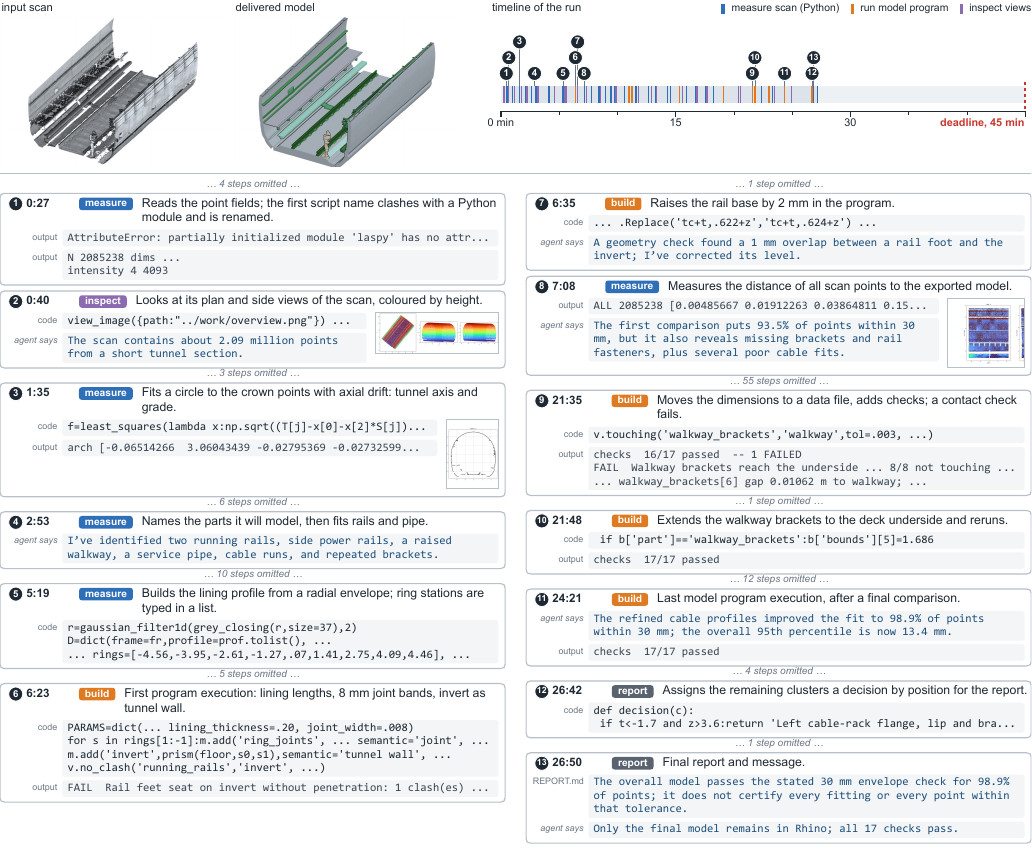}
  \caption{\textbf{GPT-6.1 Sol on STSD \texttt{horse-1-6\_2} (tunnel).} The run takes 27 minutes and executes the model program 9 times. The delivery has 276 components, lies within 5\,cm of 99.7\% of the scan points, and reaches 47.3\% mIoU. The agent fits the tunnel axis (step 3) and a lining profile. It enters the stations of the ring joints as a list of constants (step 5) and models the joints as bands 8\,mm wide (step 6). It assigns the invert to the tunnel wall (step 6). Two declared checks fail and are repaired in the geometry: an interference between rail and invert (steps 6--7) and the contact of the walkway brackets (steps 9--10). The remaining clusters of unexplained points receive a decision by position for the report (step 12).}
  \label{fig:traj_sol}
\end{figure}

\paragraph{Recorded runs.}
\Cref{fig:traj_opus,fig:traj_kimi,fig:traj_sol} show abridged timelines of three of the sampled runs, on an interior, a bridge, and a tunnel. Each figure shows the input scan, the delivered model, and a timeline with one tick per tool call (minutes from the start of the run, colored as in \cref{fig:method}). Numbered markers locate the steps shown below the timeline, and the number of steps omitted between them is given. For each step, the first line is our one-sentence description. The fields below it are verbatim from the run record, with omissions marked by dots: code that the agent ran, its output, the agent's reasoning where the harness records it, and its messages. For the Sol run, the harness records only the titles of the reasoning steps, so no reasoning is shown. Thumbnails are images that the agent viewed at that step. They are either plots produced by the agent's own code, whose colors encode its own measurements such as height or distance to its model and never labels, or views of the model rendered by the toolkit. Evaluation scores were computed after the run and were not available to the agent.

\section{Qualitative Examples}
\label{sec:supp_gallery}
\begin{figure*}[p]
  \centering
  \includegraphics[width=\linewidth]{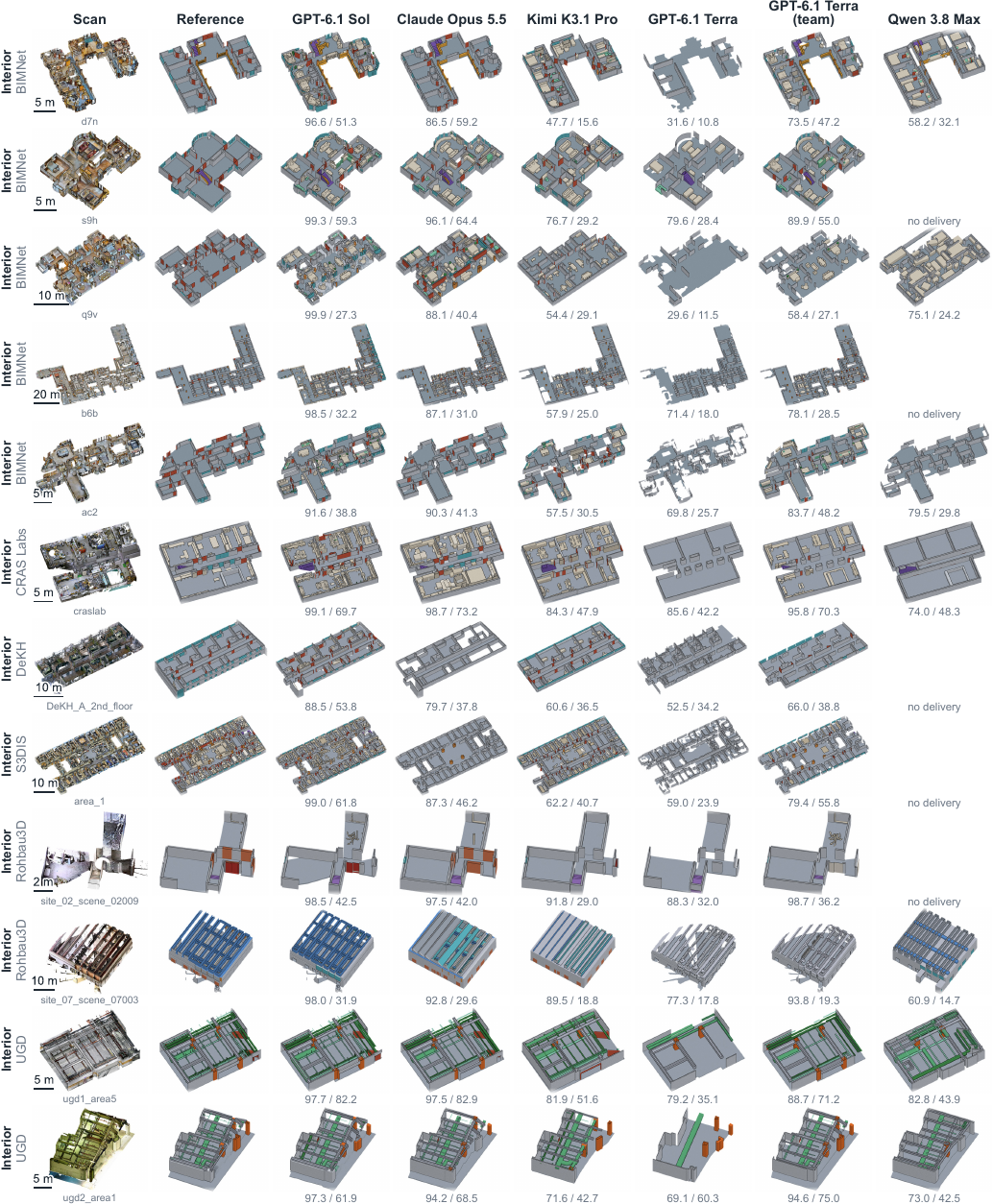}
  \caption{\textbf{Qualitative examples: interiors (1/2).} Layout and colors as in \cref{fig:qualitative}. Numbers under each model are R5 / mIoU (\%).}
  \label{fig:gallery_interior_a}
\end{figure*}

\begin{figure*}[p]
  \centering
  \includegraphics[width=\linewidth]{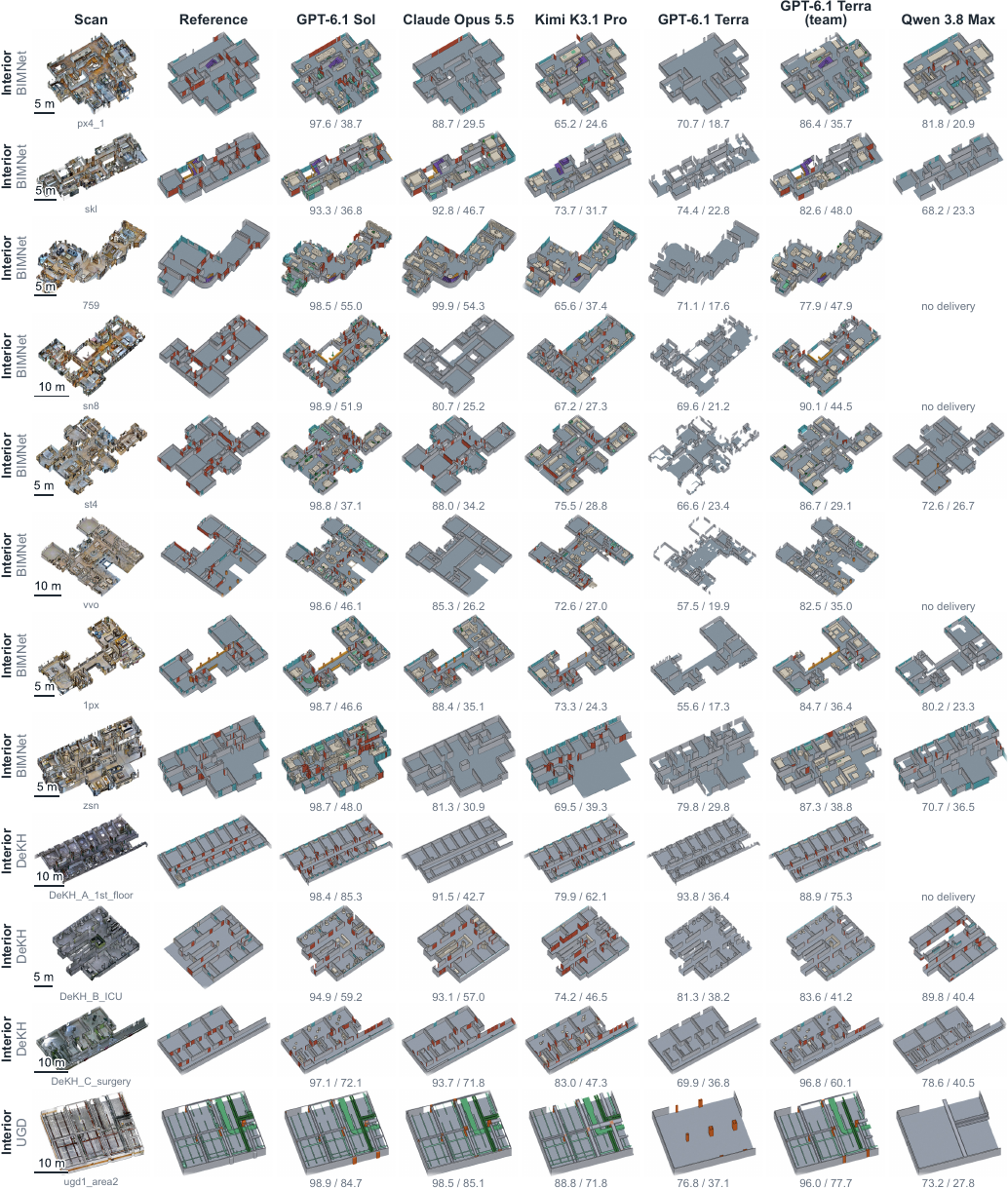}
  \caption{\textbf{Qualitative examples: interiors (2/2).} Layout and colors as in \cref{fig:qualitative}. Numbers under each model are R5 / mIoU (\%).}
  \label{fig:gallery_interior_b}
\end{figure*}

\begin{figure*}[p]
  \centering
  \includegraphics[width=\linewidth]{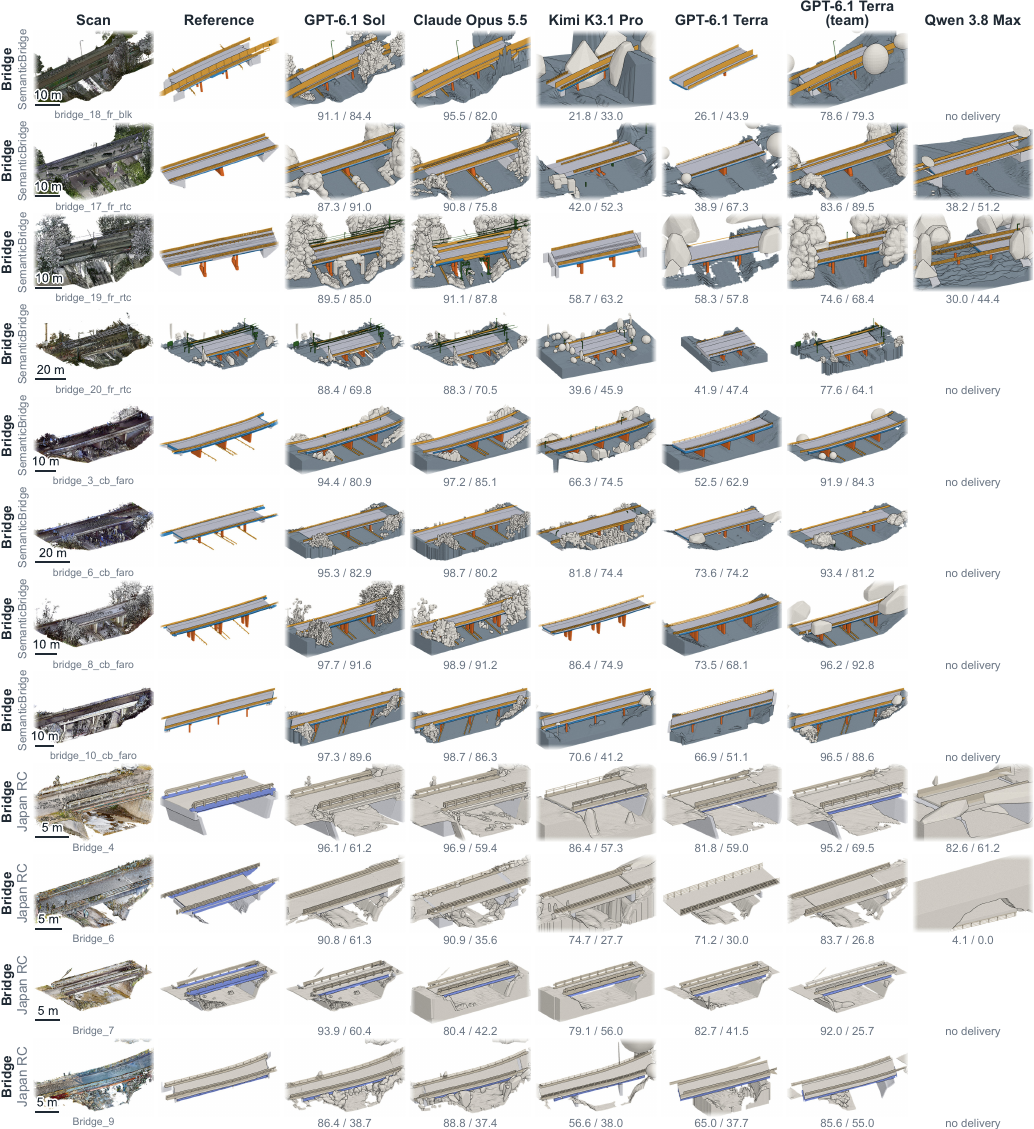}
  \caption{\textbf{Qualitative examples: bridges.} Layout and colors as in \cref{fig:qualitative}. Numbers under each model are R5 / mIoU (\%).}
  \label{fig:gallery_bridge}
\end{figure*}

\begin{figure*}[p]
  \centering
  \includegraphics[width=\linewidth]{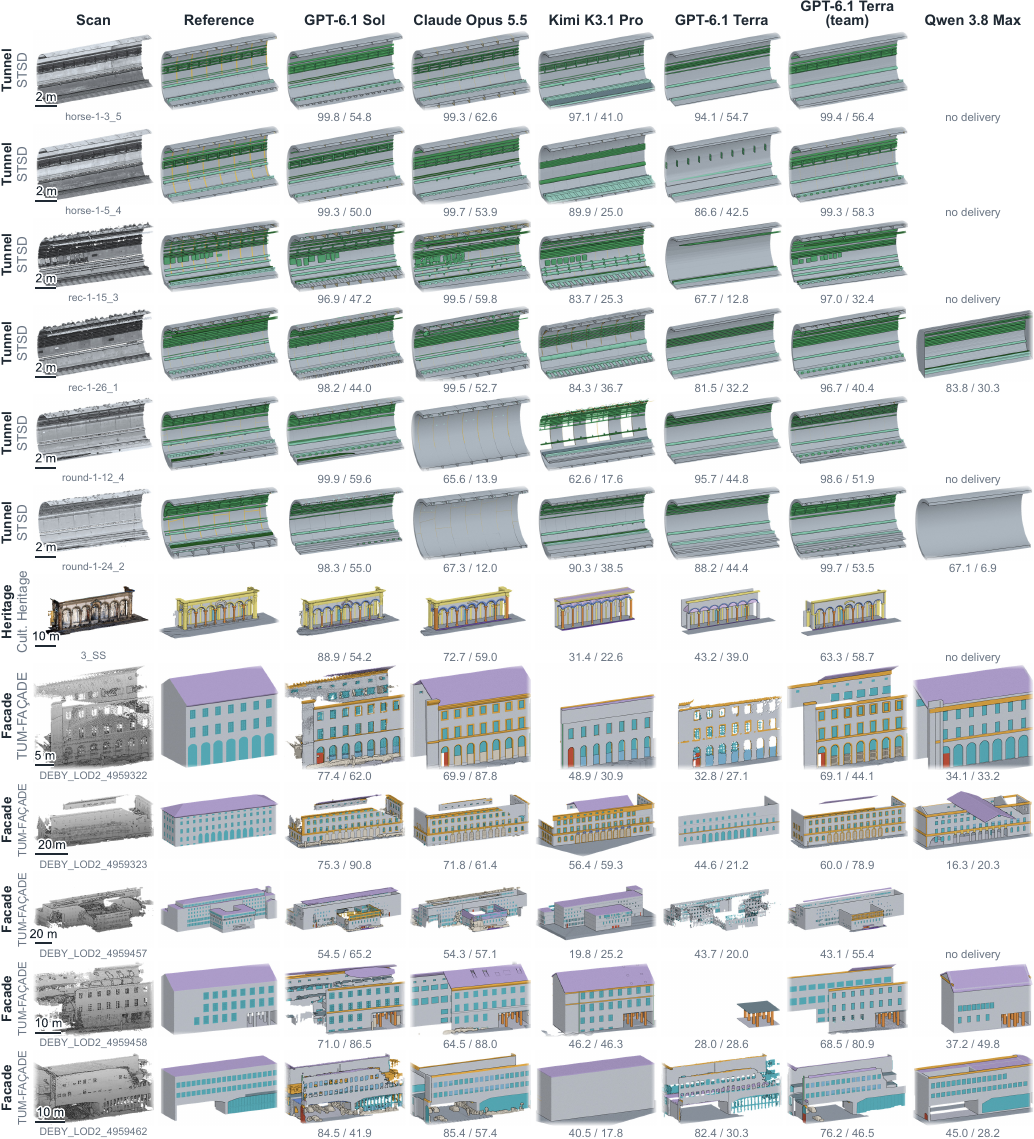}
  \caption{\textbf{Qualitative examples: tunnels, heritage, and facades.} Layout and colors follow \cref{fig:qualitative}. Numbers under each model are R5 / mIoU (\%).}
  \label{fig:gallery_mixed}
\end{figure*}

\Cref{fig:gallery_interior_a,fig:gallery_interior_b,fig:gallery_bridge,fig:gallery_mixed} extend \cref{fig:qualitative} to 48 further scenes: 24 interiors, 12 bridges, six tunnels, one heritage scene, and five facades. Scene names are given under each scan, and layout and colors follow \cref{fig:qualitative}. Interiors with exposed services are cut just below the ceiling, and bridge and facade cells are framed on the reference, so surrounding scan and terrain fade out at the cell edges. Under each model we give its R5 / mIoU (\%) as scored in \cref{tab:main}, and \emph{no delivery} marks a failed run.

\end{document}